%% file: main.tex
\documentclass[11pt]{article}
\usepackage[a4paper,margin=2.5cm]{geometry}
\usepackage[authoryear,round,longnamesfirst]{natbib}
\usepackage{amsmath,amssymb,amsfonts}
\usepackage{algorithmic}
\usepackage{graphicx}
\usepackage{textcomp}
\usepackage{tikz}
\usepackage{multirow}
\usepackage{tabularx}
\usepackage[table]{xcolor}
\usepackage{url}
\usepackage{todonotes}
\usepackage{enumitem}
\usepackage[caption=false,font=footnotesize]{subfig}
\usepackage{authblk}
\usepackage{microtype}
\newenvironment{keywords}{\par\noindent\textbf{Keywords---}}{\par}

\begin{document}

\title{Impact of Patient Orientation in Single- and Multi-View Camera Environments for AI-based Rehabilitation Monitoring}

\author[1]{Miriama~J\'{a}no\v{s}ov\'{a}\thanks{Corresponding author. E-mail: m.janosova@mail.muni.cz. ORCID: 0009-0008-4807-1023.}}
\author[2]{Andreas~Lang\thanks{E-mail: andreas.lang@tu-dortmund.de. ORCID: 0000-0003-3212-5548.}}
\author[3]{Petra~Budikova\thanks{E-mail: budikova@visioncraft.cz. ORCID: 0000-0003-1523-1744.}}
\author[1]{Jan~Sedmidubsky\thanks{Corresponding author. E-mail: sedmidubsky@mail.muni.cz. ORCID: 0000-0002-7668-8521.}}

\affil[1]{Faculty of Informatics, Masaryk University, Brno, Czech Republic}
\affil[2]{TU Dortmund University, Dortmund, Germany}
\affil[3]{VisionCraft, Brno, Czech Republic}

\date{Manuscript submitted for review on August 7, 2026.}

\maketitle

\begin{abstract}
Automated quality assessment of rehabilitation exercises relies heavily on accurate human pose estimation from video data. Although numerous RGB-based pose estimation methods have been proposed, the impact of camera placement on detecting clinically relevant movement errors remains insufficiently explored. To address this gap, we introduce REHAB26-ViewAngles, a dataset comprising correct and incorrect rehabilitation exercise executions captured from a wide range of camera angles. Furthermore, we propose a novel separability metric to quantify an algorithm's ability to distinguish between valid and faulty exercise repetitions. Using these tools, we analyze how various RGB-based pose-estimation strategies are suitable for exercise quality assessment under varying camera placements. In particular, we analyze single-camera 2D and 3D pose estimation and four multi-camera strategies: a combination of two orthogonal 2D views, 3D triangulation, weighted 3D fusion, and an AI-based pose-estimation transformer model specifically trained from two synchronized cameras.  Our findings reveal that an optimally placed 2D camera can improve the separability by 16.9\,\% over the commonly used $0^\circ$ frontal view and frequently outperforms single-camera 3D estimation, while combining two views can further improve accuracy by up to 13.1\,\%. These results offer practical guidance for deploying rehabilitation monitoring in both home and clinical settings.

\end{abstract}

\begin{keywords}
multi-view pose estimation, skeleton sequence, rehabilitation exercise, human body keypoint, camera viewpoint, motion features, exercise quality assessment, exercise similarity
\end{keywords}

\section{Introduction}

The demand for physiotherapy and rehabilitation is steadily increasing, driven by aging populations and the rising prevalence of musculoskeletal disorders~\citep{gimigliano2017world}. At the same time, a shortage of trained physical therapists has accelerated the transition toward home-based and minimally supervised rehabilitation. In such settings, patient improvements depend not only on whether exercises are performed, but also on how accurately they are executed, as even small deviations or compensatory movements can reduce therapeutic effectiveness~\citep{argent2018patient,zhang2022digital}.

Recent advances in AI-driven computer vision enable new forms of rehabilitation support outside traditional clinical environments, particularly in home-based and minimally supervised settings~\citep{bonato2024position, park2022real}. Modern \textit{pose estimation} methods~\citep{DD22,Sardari2023,yolo26,nogueira2025markerless} can extract \textit{skeletal representations} from standard RGB video in real time on consumer smartphones, providing positions of \emph{joints} in either 2D (measured in pixels of video) or estimated 3D (measured in virtual meters) space. Such skeleton data can be used to evaluate rehabilitation correctness, measure progress, and provide general guidance for patients during home rehabilitation.

Skeleton-based analysis of movement correctness has gained increasing attention across a variety of application domains, including sports training and fitness exercise~\citep{Zhou26}. In these settings, user's movements are typically represented as skeletal joint sequences and evaluated against a pre-trained model of the target exercise. However, physical rehabilitation presents additional challenges, as safe and effective therapy must account for each patient's individual capabilities, limitations, and recovery progress. Consequently, universal motion models are generally insufficient for rehabilitation monitoring, necessitating the use of personalized models that capture patient-specific movement patterns.

\subsection{Personalized Rehabilitation Monitoring}

\begin{figure}[!ht]
    \centering
    \includegraphics[width=\linewidth]{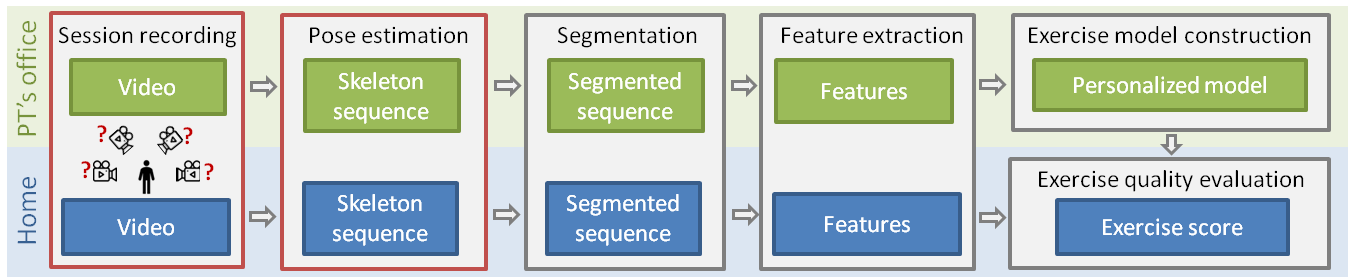}
    \caption{High-level scheme of personalized monitoring of rehabilitation exercises. The highlighted red parts are the main focus of this paper -- we are especially interested in the optimal positioning of camera(s) to capture the patient's movements.}
    \label{fig:Pipeline}
\end{figure}

In a real-world rehabilitation scenario, we cannot expect to collect many examples of correct and incorrect exercise execution that would be necessary for training a personalized neural-network (NN) model. Instead, recent personalized AI-based rehabilitation monitoring systems employ similarity-based analysis using a few examples of patient's exercise collected during supervised patient's sessions with a therapist. In this work, we adopt the exercise monitoring pipeline proposed in~\citep{Janosova26}, which is illustrated in Fig.~\ref{fig:Pipeline}.

The pipeline assumes that the patient first visits the physiotherapist (PT), who determines suitable exercise types and teaches the patient how to perform them correctly, taking into account their current needs and abilities. During the practice, the PT records several exercise \emph{repetitions} using a mobile application and provides a simple evaluation of whether the patient performed them according to instructions. From these examples, the personalized model of correct exercising is created.
As illustrated in Fig.~\ref{fig:Pipeline}, both PT-guided and home exercises are processed through the same computational pipeline, which includes video recording, skeletal pose estimation, automatic segmentation into individual repetitions, and extraction of interpretable \textit{features} that describe relationships between body segments. \emph{Correctly} performed repetitions identified by the physiotherapist are used to (1) estimate personalized \emph{confidence intervals} that capture acceptable variation in individual feature values over time, and to (2) select a single repetition as a \textit{personalized reference}.

During home exercise, each repetition is temporally aligned with the reference and evaluated by \emph{similarity analysis}, which compares the feature values in individual frames to their respective confidence intervals. Feature weighting based on discriminative importance further enables the computation of an overall quality score and the identification of the most relevant movement errors~\citep{JBS24}. A more detailed explanation of the exercise quality analysis will be provided in Section~\ref{sec:Formalisms}.

\subsection{Patient Orientation Challenge}

Accurate assessment of rehabilitation exercises depends strongly on the quality of skeleton data, which in turn depends on camera placement. Unlike the dynamic, full-body movements commonly studied in SOTA pose estimation research, physiotherapy exercises typically involve slow, controlled motions with limited ranges of movement. In this context, pose estimation errors of only a few degrees may be clinically significant. Joint visibility and self-occlusion can change substantially across viewpoints, directly affecting the reliability of movement analysis.

Camera placement is further constrained by practical considerations. In home-based rehabilitation, patients typically rely on a single, uncalibrated camera, making viewpoint selection especially critical. However, if the expected gains in accuracy are substantial, the use of two smartphones to record exercises in both clinical and home settings may be feasible. The use of more than two cameras is considered unlikely.

In this paper, we study how \textit{camera viewpoint} influences pose estimation and exercise-quality assessment in rehabilitation using similarity-based data analysis. We evaluate the effect of camera viewing angle for common physiotherapy exercises, focusing on real-world settings where personalized exercise models need to be constructed from limited training data. We evaluate single-camera configurations across a range of viewing angles, analyze robustness to small changes in patient orientation, and compare the suitability of 2D and 3D pose representations obtained from monocular pose estimation models. Furthermore, we explore several strategies for using multiple uncalibrated cameras, including a combination of two orthogonal 2D views, geometric triangulation, merging of independently estimated 3D skeleton sequences, and the cross-view transformer-based RUMPL framework~\citep{ghasemzadeh2026rumpl} trained on multi-view inputs.

The main contributions of this paper are as follows:
\begin{itemize}
    \item We design and collect a new dataset (REHAB26-ViewAngles) comprising typical rehabilitation exercises recorded from a wide range of angles. The dataset contains annotated examples of both correct and incorrect exercises and enables assessing the impact of a person's orientation towards camera(s) on the quality of exercising.
    \item We establish an automated evaluation methodology using personalized models of correct exercise execution, which are essential for clinically relevant monitoring. This methodology features a novel separability metric that quantifies an algorithm’s ability to distinguish between correct and incorrect repetitions.
    \item We investigate 2D and 3D pose estimation strategies, including monocular, multi-view, and learning-based cross-view transformer (RUMPL) approaches, to obtain robust skeletal representations from RGB video.
    \item We experimentally evaluate approaches to single- and multi-camera skeleton extraction over the new dataset. For selected SOTA pose estimation methods, we identify the optimum patient orientations.
\end{itemize}

The remainder of this paper is organized as follows. Section~2 reviews the related work in pose estimation and skeleton-based rehabilitation monitoring. Section~3 introduces the proposed REHAB26-ViewAngles dataset~\citep{janosova2026rehab26}. Section~4 explains the new separability measure, presents investigated approaches to skeleton extraction using single- and multi-camera setups, and formulates the research questions addressed in the experimental evaluation. Section~5 presents and discusses the experimental evaluation, resulting in practical recommendations for camera placement across different rehabilitation exercises. Finally, Section~6 concludes the paper and outlines directions for future work.

\section{Related Work}

Reliable AI-based rehabilitation monitoring depends on several interconnected research areas. This section reviews recent advances in human pose estimation, skeleton-based exercise analysis, and studies addressing the impact of camera viewpoint on pose estimation and motion analysis. Together, these topics provide the foundation for the research questions investigated in this paper.

\textbf{Pose Estimation Methods.}
Skeleton data are most often obtained from RGB video by applying human pose estimation (HPE) methods that extract joint locations from each video frame \citep{Debnath2022,Sardari2023}.
2D pose estimation aims to localize body joints in the image plane and serves as the foundation for many 3D approaches. Most monocular 3D pose estimation methods follow a 2D-to-3D paradigm, where 2D joint locations are detected and subsequently lifted to 3D space~\citep{akhter2015pose,ramakrishna2012reconstructing}, though some approaches estimate 3D poses directly via volumetric heatmaps~\citep{pavlakos2017coarse}. More recently, learning-based methods have improved 2D-to-3D regression using transformers~\citep{li2022mhformer,tang20233d} or graph-based models~\citep{liu2021graph,zhao2022graformer} to encode skeletal relationships. While these designs reduce depth ambiguity, they remain fundamentally constrained by the limitations of a single viewpoint, leading recent studies~\citep{sun2025depth, vcernek2025pose} to suggest that 2D estimation often yields more accurate and stable joint localization overall.
On the other hand, the 2D skeleton is often deformed by the camera viewpoint and cannot reliably capture out-of-plane motion or joint angles~\citep{klein2023assessing}.

Multi-view approaches address some of these limitations by exploiting synchronized camera feeds, either through geometric triangulation of 2D detections~\citep{hartley1997triangulation} or by learning to fuse multi-view features directly in 3D space~\citep{tu2020voxelpose,zhang2021direct,li2025mvgformer,chharia2025mv}.  However, standard multi-view networks often rely on fixed camera configurations, limiting their adaptability in home environments. To address this, recent frameworks like RUMPL~\citep{ghasemzadeh2026rumpl} map 2D keypoints into 3D ray vectors, eliminating the dependency on rigid camera arrangements. While this formulation allows the system to adapt to random camera placements, the resulting 3D reconstruction still depends on valid camera calibration to orient the vectors. Consequently, the performance of such models remains largely unexamined for capturing the subtle clinical movements required for rehabilitation, especially when limited to sparse, two-camera arrays.

Given the spatial and computational requirements of multi-camera systems, practical rehabilitation monitoring typically relies on off-the-shelf monocular pose estimation solutions, with MediaPipe Pose (MPP)~\citep{bazarevsky2020blazepose} being among the most widely adopted. Owing to its real-time performance and efficient deployment on consumer mobile devices~\citep{vcernek2025pose}, it is particularly well suited for home-based rehabilitation applications. MediaPipe Pose combines a BlazePose-based 2D pose detector with the GHUM body model~\citep{xu2020ghum} to estimate both 2D and 3D skeletal poses directly from monocular RGB video, providing a favorable balance between accuracy, computational efficiency, and ease of deployment.

\textbf{Exercise Correctness Analysis.}
In rehabilitation, the primary objective is to determine whether an exercise is performed correctly while identifying and explaining movement errors. At the same time, rehabilitation monitoring systems should be readily accessible to the general public, making solutions based on standard consumer devices particularly attractive. Although some approaches rely on wearable sensors~\citep{ lu2020evaluating}, recent research has increasingly focused on vision-based methods that operate on monocular RGB video acquired using standard mobile devices. These methods typically employ human pose estimation followed by feature extraction to transform high-dimensional movement into clinically meaningful descriptors~\citep{williams2019assessment}. Practical implementations have recently demonstrated that this entire extraction pipeline can operate in real time directly on consumer smartphones~\citep{janosova_demo}.

Inspired by advances in general human motion analysis, some approaches learn generic representations of correct exercise execution using deep neural networks~\citep{HCWC24}. However, such models generally require large amounts of annotated training data and are therefore difficult to personalize for individual patients in practical rehabilitation settings. Consequently, recent personalized rehabilitation monitoring systems predominantly rely on interpretable kinematic features derived from skeletal pose sequences, such as joint angles, limb orientations, or relative joint positions over time~\citep{mccay2021pose,MR06}. During assessment, the feature trajectories extracted from a patient's exercise are compared with those of a personalized reference using temporal alignment and similarity measures such as Dynamic Time Warping (DTW)~\citep{PCVLMS24}. Recent work has further improved this framework by introducing individualized feature weighting, allowing clinically more relevant movement characteristics to have a greater influence on the overall assessment~\citep{JBS24}.

\textbf{Role of Camera Viewpoint.}
Compared with advances in human pose estimation and exercise quality assessment, the influence of camera viewpoint has received relatively little attention. Most existing rehabilitation monitoring systems assume a fixed camera placement, typically using a frontal or semi-profile view, without systematically evaluating whether alternative viewpoints provide more reliable movement information. However, camera viewpoint directly affects joint visibility, self-occlusion, and ultimately the accuracy of pose estimation. Consequently, the ability to detect clinically relevant movement errors also depends on the chosen viewpoint. For example, squat depth and knee alignment are more readily observed from lateral views, whereas frontal views better capture arm symmetry and shoulder positioning.

Recent studies provide growing evidence that viewpoint substantially influences the quality of skeleton-based movement analysis. \citet{klein2023assessing} demonstrated that MediaPipe Pose produces markedly different joint angle estimates across camera viewpoints, with the largest errors occurring when the upper limbs are oriented toward the camera. Similarly, \citet{baldinger2025influence} reported that posterior viewpoints yield more stable lower-limb measurements owing to reduced self-occlusion, while accurate shoulder estimation remains challenging regardless of viewpoint. Furthermore, \citet{zollner2025evaluation} showed that camera angle strongly affects joint visibility during dynamic exercises, where changes in body orientation introduce additional viewpoint variability.

Despite these findings, a comprehensive evaluation of viewpoint selection for rehabilitation monitoring is still lacking, largely because suitable datasets are unavailable. Existing datasets for both general human motion analysis~\citep{Human3.6Dataset, NTUDataset, MPI-INF-3DHPDataset} and rehabilitation exercises~\citep{FineRehabDataset,capecci2019kimore,CSB24-REHAB24-6,vakanski2018data} are typically recorded from only a small number of predefined viewpoints, most commonly frontal, lateral, or semi-profile views. Consequently, they do not enable a systematic investigation of patient orientation or camera placement, motivating the creation of the REHAB26-ViewAngles dataset presented in this work.

\section{REHAB26-ViewAngles Rehabilitation Dataset}

\begin{figure}[!t]
    \centering
    \includegraphics[width=0.95\linewidth]{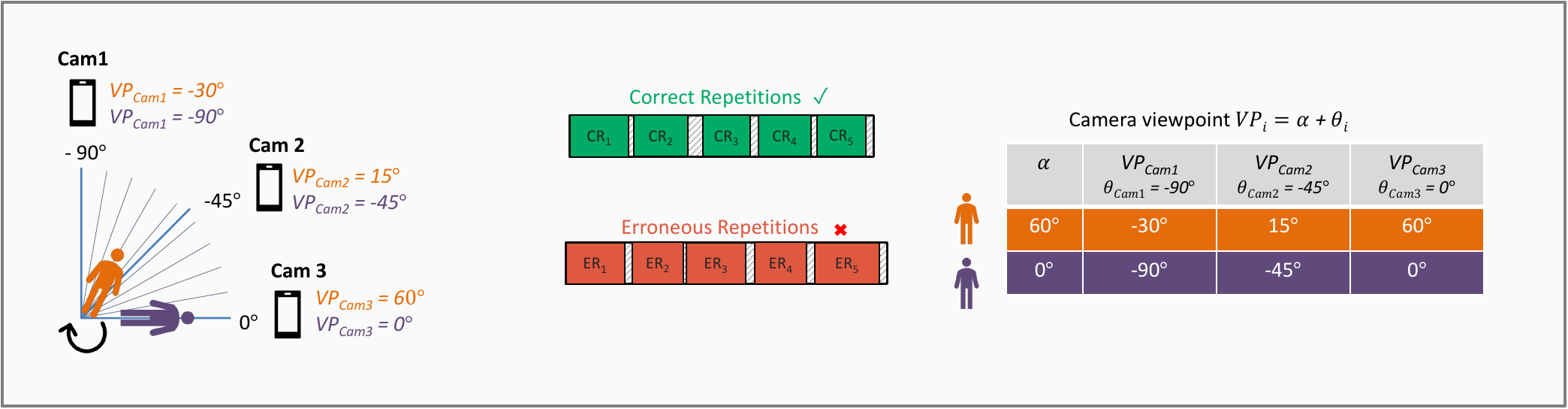}
    \caption{The REHAB26-ViewAngles data acquisition setup. Three cameras are positioned at fixed angles ($\theta_{Cam1} = -90^\circ$, $\theta_{Cam2} = -45^\circ$, $\theta_{Cam3} = 0^\circ$) while the participant systematically changes their standing orientation (angle $\alpha$ with respect to Cam3). At each designated orientation, the participant performs a continuous set of 10 repetitions, consisting of 5 correct and 5 erroneous repetitions, each with a specific clinical error. The camera viewpoint ($VP_{Cam_i}$) for any given recording is mathematically defined as the sum of the participant's rotation and the static camera position ($VP_{Cam_i} = \alpha + \theta_i$).}
    \label{fig:datasetRecording}
\end{figure}

To enable a systematic study of camera viewpoint in rehabilitation monitoring, we collected a new dataset, REHAB26-ViewAngles~\citep{janosova2026rehab26}, specifically designed to evaluate the influence of patient orientation on pose estimation and exercise quality assessment. Unlike existing rehabilitation datasets, which primarily aim to benchmark exercise recognition or movement assessment from a small number of predefined viewpoints, REHAB26-ViewAngles was designed around several considerations motivated by practical home-based rehabilitation.

First, rather than maximizing the number of exercise types, we selected a representative set of rehabilitation exercises covering the most common movement patterns, body regions, and directions of motion encountered in clinical practice. This design allows us to investigate viewpoint effects across diverse movement characteristics while keeping the dataset sufficiently controlled.

Second, the dataset emphasizes viewpoint diversity. Each exercise was recorded from a dense set of viewing angles covering the frontal hemisphere, enabling a detailed analysis of how patient orientation affects pose estimation and subsequent movement assessment. In addition to the nominal recording angles, small orientation offsets were intentionally allowed to simulate realistic situations in which patients do not position themselves exactly as instructed before starting an exercise.

Third, each exercise was performed both correctly and with clinically relevant execution errors identified by experienced physiotherapists. This enables not only the evaluation of pose estimation accuracy, but also the assessment of whether different viewpoints allow clinically meaningful movement deviations to be reliably detected.

Finally, the recording setup reflects the intended deployment scenario. All videos were captured simultaneously using identical consumer smartphones, allowing both single-camera and multi-camera evaluation under realistic conditions while eliminating hardware-induced variability. Simultaneous recordings further provide temporally synchronized multi-view sequences suitable for evaluating multi-camera pose estimation and skeleton fusion methods. To ensure precise frame alignment across devices, temporal synchronization was established visually by recording a digital timer displaying milliseconds on a tablet visible from all viewpoints.

\subsection{Exercise Selection and Clinical Error Modeling}

The REHAB26-ViewAngles dataset comprises three representative physiotherapy exercises selected to cover different body regions, movement directions, and biomechanical characteristics commonly encountered in rehabilitation:

\begin{itemize}
    \item \textbf{Exercise 1 (Arm Abduction):} Lateral elevation of the right arm from a neutral position to the maximum position, followed by a return to the starting position.
    \item \textbf{Exercise 2 (Shoulder Extension):} Posterior movement of the right arm behind the body from a neutral position, followed by a return.
    \item \textbf{Exercise 3 (Squat):} Lowering the body by flexing the hips and knees while maintaining an upright torso, followed by a return to the standing position.
\end{itemize}

\begin{table*}[!t]
\centering
\caption{Overview of movement errors, their associated joint-angle measurements, and the corresponding feature codes. Some features (F2, F4-F6) are shared across multiple exercises. Features F9–F14 are defined separately for the left and right sides.}
\begin{tabularx}{\textwidth}{l|X|c|X}
\textbf{Exercise} & \textbf{Error Description} & \textbf{Feature Code} & \textbf{What Angle is Measured?} \\
\hline
Ex1 & Arm positioned too far forward & F1 & Angle between the hand, shoulder, and torso \\
Ex1, Ex2 & Shoulder elevation & F2 & Angle between both shoulders, and hip\\
Ex1 & Lateral head tilt & F3 & Angle between the head, the torso, and vertical axis \\
Ex1, Ex2 & Elbow flexion (frontal) & F4 & Angle between the upper arm and forearm \\
Ex1, Ex2 & Wrong arm positioning & F5 & Angle between the hand and both shoulders \\
Ex2, Ex3 & Lateral torso bending & F6 & Angle between the torso and the lower limbs \\
Ex2 & Torso rotation & F7 & Angle between the head, torso, and lower body\\
Ex2 & Elbow flexion (lateral) & F8 & Angle between the upper arm and forearm \\
Ex3 & Knees angle inward & F9, F10 & Angle between thigh, knee, and toes \\
Ex3 & Knees pass toes & F11, F12 & Angle between toe, knee, and vertical axis \\
Ex3 & Lifted heels & F13, F14 & Angle between hip, heel, and toe \\
Ex3 & Insufficient squat depth & F15 & Angle between hip, knee, and ankle  \\
\hline
\end{tabularx}
\label{tab:features}
\end{table*}

Rather than maximizing the number of exercise types, we intentionally selected a small yet representative set of movements exhibiting different kinematic properties. Together, the exercises cover both upper- and lower-body rehabilitation, involve motion in different anatomical planes, and require substantially different camera viewpoints for reliable observation. For example, shoulder abduction is most naturally observed from a frontal view, whereas shoulder extension and squat depth are better captured from lateral viewpoints. The visual and spatial characteristics of the selected exercises are illustrated in Fig.~\ref{fig:dataset_examples}.

To support the evaluation of clinically meaningful exercise monitoring methods rather than pose estimation alone, each exercise was additionally performed with five common execution errors defined in collaboration with experienced physiotherapists (Table~\ref{tab:features}). Examples include knee valgus, excessive trunk flexion, and insufficient range of motion. Each error was associated with a dedicated interpretable feature derived from the skeletal representation, enabling the evaluation of whether different camera viewpoints preserve the information required for reliable error detection.

\subsection{Acquisition Protocol}
\label{sec:AcquisitionProtocol}

\begin{figure}[!t]
    \centering
    \subfloat[]{\includegraphics[width=0.5\linewidth]{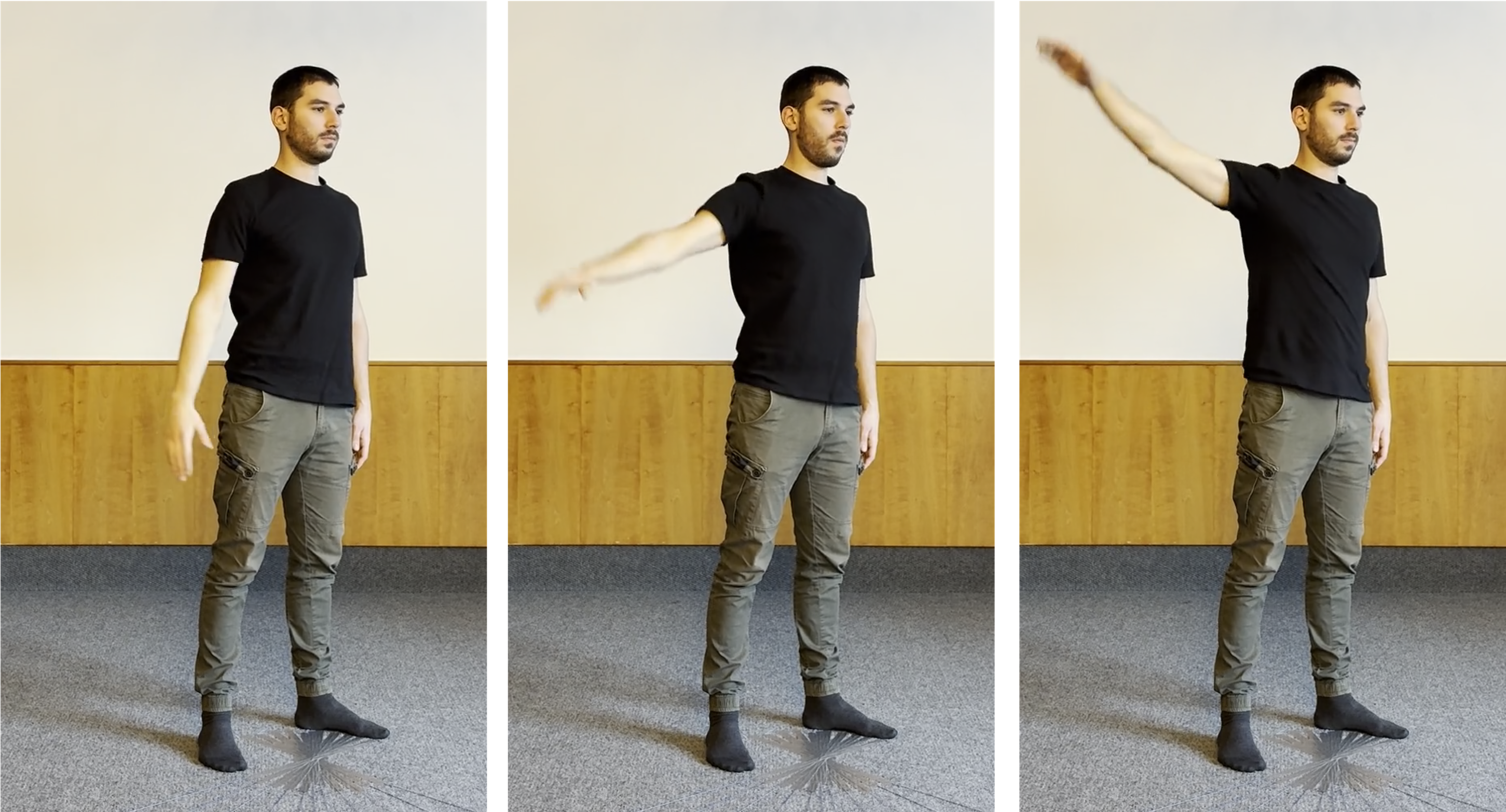}\label{fig:ex_progression}} \\
    \subfloat[]{\includegraphics[width=0.5\linewidth]{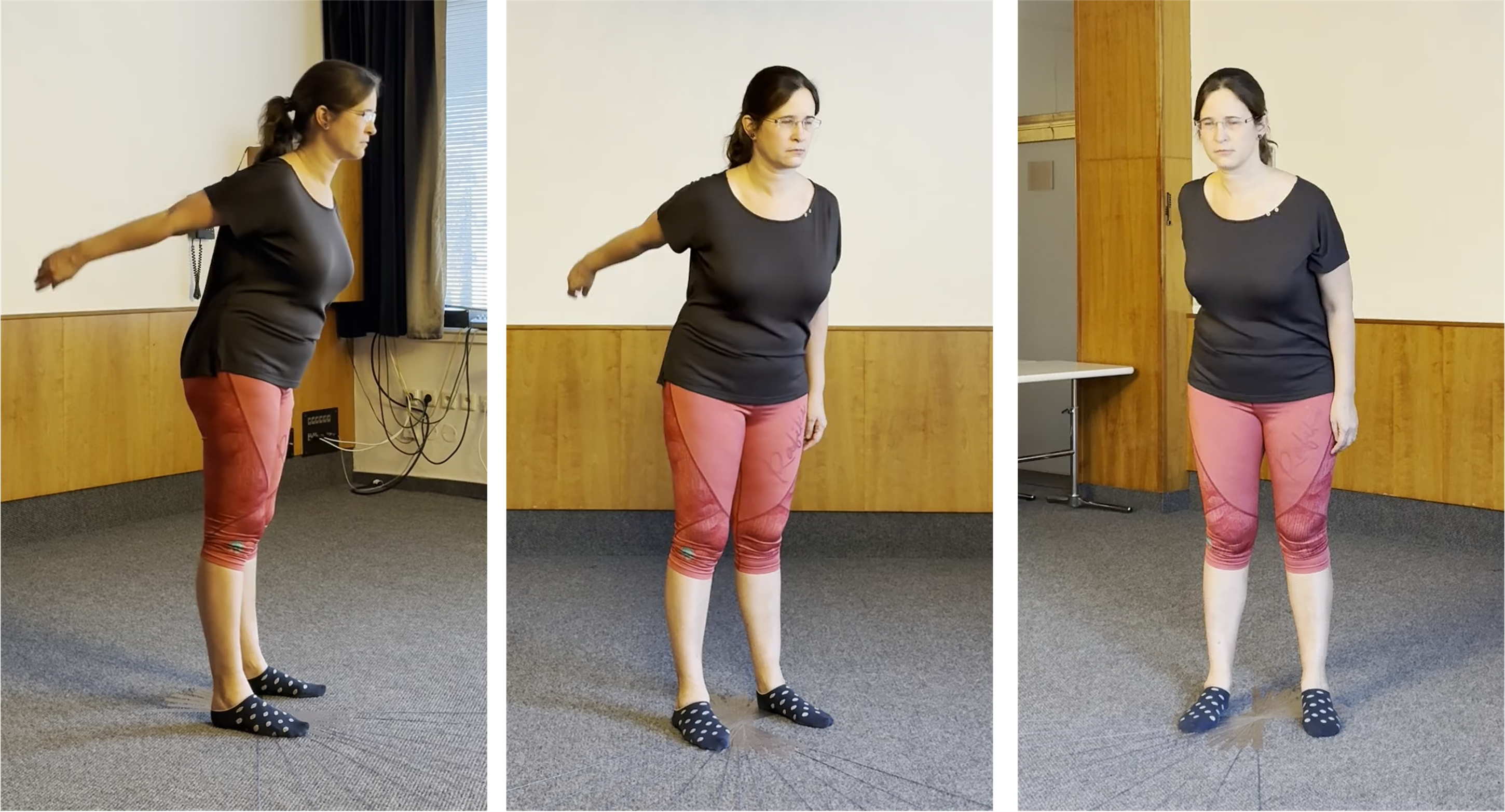}\label{fig:ex_multicam}}\\
    \subfloat[]{\includegraphics[width=0.5\linewidth]{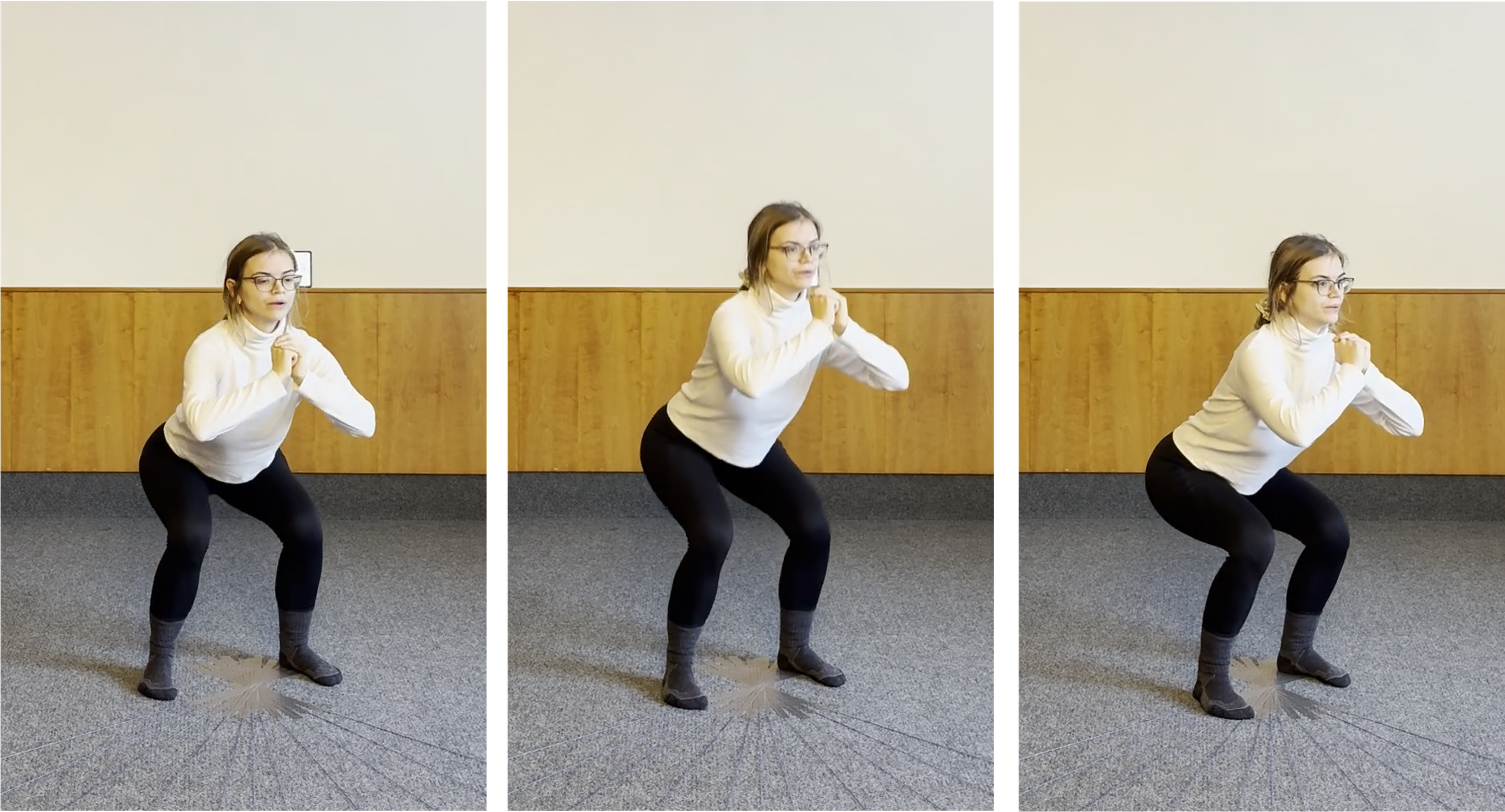}\label{fig:ex_angleshift}}
    \caption{Representative examples from the REHAB26-ViewAngles dataset. (a) Temporal progression of Exercise 1 (Arm Abduction) recorded from a single camera. (b) Exercise 2 (Shoulder Extension) recorded simultaneously from three distinct camera perspectives.
    (c) Exercise 3 (Squat) captured from the same camera across three different patient orientations, with the viewpoint shifting by 10-degree increments. }
    \label{fig:dataset_examples}
\end{figure}

To record the exercise, we utilized three synchronized iPhone 16e smartphones (30~FPS, 12~MP, $f/1.6$) positioned in a fixed horizontal arrangement, as illustrated in Fig.~\ref{fig:datasetRecording}. To capture viewpoint variability, participants performed exercise sets while systematically shifting their orientation using a large floor protractor. A patient-centric coordinate system was adopted, where ($0^\circ$) corresponds to a frontal view, positive angles denote rotations to the right, and negative angles rotations to the left. Participants initially faced Cam1 and subsequently rotated toward Cam3 in ($10^\circ$) increments between exercise sets. Cam1, Cam2, and Cam3 were placed at fixed relative positions of $-90^\circ$, $-45^\circ$, and $0^\circ$ with respect to the participant facing Cam3. As illustrated in Fig.~\ref{fig:datasetRecording}, this protocol produces relative viewing angles spanning the frontal hemisphere, from a $+90^\circ$ right profile to a $-90^\circ$ left profile. An additional reference recording was acquired with participants directly facing Cam2. Several examples of exercise recorded from various angles are shown in Fig.~\ref{fig:dataset_examples}.

Five volunteers (two male and three female) with different body types and fitness levels participated in the data collection. For every exercise and nominal viewing angle, each participant performed five correct repetitions together with five repetitions containing predefined execution errors, one for each clinically relevant error type. To avoid recording multiple nearly identical repetitions, correct and incorrect executions were intentionally interleaved rather than grouped together. Our previous experience with rehabilitation data collection showed that consecutive correct repetitions tend to become unnaturally consistent, whereas alternating correct and incorrect executions introduces the natural variability observed during home exercise. All recordings were manually segmented into individual repetitions and annotated by the supervising physiotherapist.

Although participants were instructed to align themselves with the prescribed viewing angles, small deviations in body orientation were intentionally preserved. Consequently, the ($10^\circ$) increments represent nominal rather than exact viewing angles. This variability closely reflects realistic home-rehabilitation scenarios, where patients are unlikely to position themselves with perfect accuracy relative to the camera. Preserving these naturally occurring orientation offsets therefore enables the evaluation of viewpoint robustness under practical deployment conditions.

\subsection{Dataset Statistics}

The resulting REHAB26-ViewAngles dataset comprises 3 rehabilitation exercise types performed by 5 participants under 11 nominal body orientations. For every combination of exercise, participant, and orientation, 10 repetitions were recorded, including 5 correct executions and 5 incorrect executions corresponding to predefined clinical error types. In total, the dataset contains 1,650 manually annotated exercise repetitions, evenly divided between correct and incorrect execution. Each erroneous repetition is additionally labeled with the specific type of clinical error performed.

All repetitions were recorded simultaneously using three synchronized cameras, resulting in observations from 29 unique viewing angles across the complete acquisition protocol. Overall, the dataset contains more than 6 hours of RGB video, comprising approximately 650,000 frames. The synchronized multi-view recordings, together with the dense sampling of patient orientations, make REHAB26-ViewAngles particularly suitable for systematically evaluating the influence of camera viewpoint on pose estimation and personalized rehabilitation monitoring.

\clearpage
\begin{table*}[!t]
\centering
\caption{Overview of all viewpoints available in REHAB26-ViewAngles dataset. A check mark indicates that the viewpoint (column header) was recorded for the given camera.}
\label{tab:dataset_angles}
\setlength\tabcolsep{2.5pt}
\footnotesize
\begin{tabular}{@{}l|*{29}{c}@{}}
\hline
 & \rotatebox{90}{$-90^\circ$} & \rotatebox{90}{$-80^\circ$} & \rotatebox{90}{$-70^\circ$} & \rotatebox{90}{$-60^\circ$} & \rotatebox{90}{$-50^\circ$} & \rotatebox{90}{$-45^\circ$} & \rotatebox{90}{$-40^\circ$} & \rotatebox{90}{$-35^\circ$} & \rotatebox{90}{$-30^\circ$} & \rotatebox{90}{$-25^\circ$} & \rotatebox{90}{$-20^\circ$} & \rotatebox{90}{$-15^\circ$} & \rotatebox{90}{$-10^\circ$} & \rotatebox{90}{$-5^\circ$} & \rotatebox{90}{$0^\circ$} & \rotatebox{90}{$5^\circ$} & \rotatebox{90}{$10^\circ$} & \rotatebox{90}{$15^\circ$} & \rotatebox{90}{$20^\circ$} & \rotatebox{90}{$25^\circ$} & \rotatebox{90}{$30^\circ$} & \rotatebox{90}{$35^\circ$} & \rotatebox{90}{$40^\circ$} & \rotatebox{90}{$45^\circ$} & \rotatebox{90}{$50^\circ$} & \rotatebox{90}{$60^\circ$} & \rotatebox{90}{$70^\circ$} & \rotatebox{90}{$80^\circ$} & \rotatebox{90}{$90^\circ$} \\
\hline
\textbf{Cam1} & \checkmark & \checkmark & \checkmark & \checkmark & \checkmark & \checkmark & \checkmark & & \checkmark & & \checkmark & & \checkmark & & \checkmark & & & & & & & & & & & & & & \\
\textbf{Cam2} & & & & & & \checkmark & & \checkmark & & \checkmark & & \checkmark & & \checkmark & \checkmark & \checkmark & & \checkmark & & \checkmark & & \checkmark & & \checkmark & & & & & \\
\textbf{Cam3} & & & & & & & & & & & & & & & \checkmark & & \checkmark & & \checkmark & & \checkmark & & \checkmark & \checkmark & \checkmark & \checkmark & \checkmark & \checkmark & \checkmark \\
\hline
\end{tabular}
\end{table*}

\section{Viewpoint Optimization for Rehabilitation Monitoring}

Using the collected REHAB26-ViewAngles dataset, we investigate how camera placement and pose estimation strategy influence the quality of personalized rehabilitation monitoring. We consider two practical deployment scenarios: (1) typical home environments, where only a single camera is available, and (2) advanced clinical or home settings employing two uncalibrated cameras. For each scenario, our objective is to identify the optimal camera placement and the most suitable approach for obtaining reliable skeletal representations from the available RGB video stream(s).

This section first introduces the exercise quality assessment framework used throughout the paper. It then provides an overview of the considered single- and multi-camera approaches and formulates the research questions addressed in the experimental evaluation. Finally, it describes the investigated skeleton acquisition methods, including their implementation details.

\subsection{Movement Quality Analysis}
\label{sec:Formalisms}

In skeleton-based movement analysis, human motion is represented as a temporal sequence $(P_1, \dots, P_n)$ of $n$ \emph{poses}, where each pose $P_i$ ($1 \leq i \leq n$)  consists of $k$ skeleton \emph{joints} $J_1, \ldots, J_k$. Depending on the 2D/3D format, each joint is defined either as $J_j \in \mathbb{R}^3$ in a 3D coordinate system (virtual meters), or $J_j \in \mathbb{R}^2$ in a 2D coordinate system (pixels of a video frame).

\textbf{Exercise Sequence, Repetition, Feature.}
In our research, we work with the concept of \emph{exercise sequence} $S$, which is a recording of a single exercise performed by a single actor in a specific position within the recording space. As described in the previous section, each exercise sequence is composed of 10 \emph{repetitions} that are either performed correctly or incorrectly. In this work, we assume that each repetition $R^S_i$ is already segmented from the recorded sequence. Let $Reps^{S}_{OK}$ be the set of all correct repetitions within exercise sequence $S$, and $Reps^{S}_{NOK}$ the set of all incorrect (erroneous) repetitions.
To analyze movement quality, we utilize \emph{features} that are computed over raw pose coordinates. The definitions of features $F_1, \ldots, F_{15}$ that are relevant for the REHAB26-ViewAngles exercises are provided in Table~\ref{tab:features}. Let $F_k(P_i)$ be the value of feature $F_k$ computed over pose $P_i$.

\textbf{Exercise Model.}
To allow personalized evaluation of rehabilitation exercise quality, a model of correct exercise needs to be created for each patient. In a real application scenario, examples of correct movement recorded in the PT's office are used to construct the model. We simulate this process and create \emph{exercise model} $M^S$ for $S$ using $Reps^{S}_{OK}$. First, we select one correct repetition at random as a \textit{reference repetition} $R^S_{ref}$. For all other correct repetitions $R_j \in Reps^{S}_{OK}$, we compute a temporal alignment between $R_j$ and $R^S_{ref}$ to compensate for differences in repetition length. Specifically, the Dynamic Time Warping (DTW) is used to align repetitions with respect to a single exercise-specific feature that captures the full range of motion (e.g., $F_5$ for Exercise~1).

\begin{figure}[!t]
    \centering
    \includegraphics[width=0.8\linewidth]{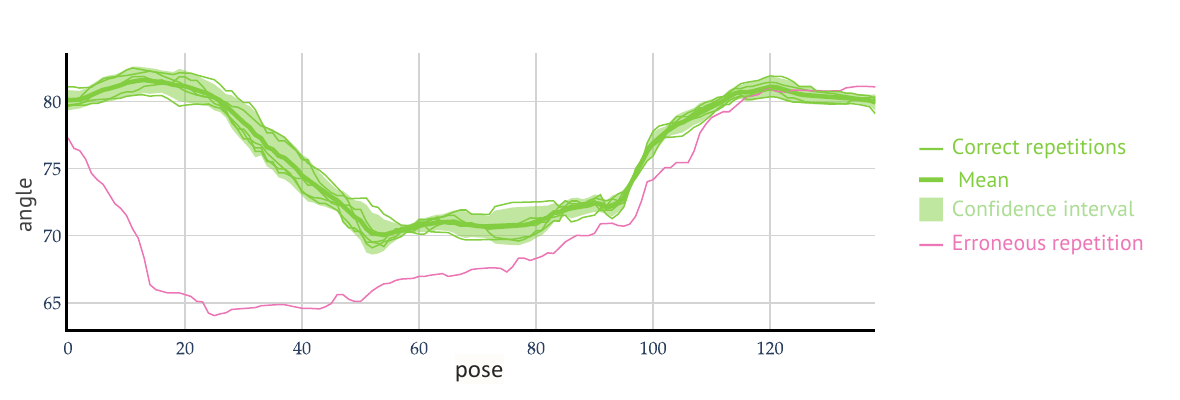}
    \caption{Comparison of a feature F2 value (angle) across Exercise1 poses for correct vs. erroneous repetitions, including the mean and 95\% confidence interval of the correct repetitions.}
    \label{fig:feature_example}
\end{figure}

After alignment, we compute the mean feature value $\mu^k_i$ and a corresponding tolerance threshold $\tau^k_i$ for each feature $F_k$ at every frame $P_i$ of $R^S_{\text{ref}}$. Here, $\tau^k_i$ is defined as the margin of error for a 95\,\% confidence interval under Student's $t$-distribution (the most suitable model for our data). This establishes the confidence interval $CI^k_i = [\mu^k_i - \tau^k_i, \, \mu^k_i + \tau^k_i]$ as the acceptable tolerance range for correct exercise execution.
To avoid overly narrow confidence intervals, we further utilize the first quartile of the values distribution as the lower bound on confidence interval width. The construction of an exercise model for a single feature is illustrated in Fig.~\ref{fig:feature_example}.

\textbf{Exercise Correctness Evaluation.}
Using the exercise model $M^S$, we can measure the correctness of any exercise repetition $R$. First, let us define a \textit{pose error} $E^k_i$ that determines the error of pose $P_i \in R$ w.r.t. feature $F_k$:
\begin{equation}
  E^k_i(R,M^S) = \begin{cases}
  0 & \quad \text{if }~\left|F_k(P_i) - \mu^{k}_i \right| \leq \tau^{k}_i\\
  min\left(1, \frac{\left|F_k(P_i) - \mu^{k}_i \right| - \tau^{k}_i}{\tau^{k}_i} \right) & \quad \text{else}.
  \end{cases}
\end{equation}
The pose error $E^k_i$ takes values in $[0, 1]$, where $0$ indicates no deviation from the reference and $1$ represents maximum deviation. To assess the entire movement sequence, the overall error $E^k$ of the complete repetition $R$ w.r.t. feature $F_k$ is computed as the average of pose errors:
\begin{equation}
  E^k(R,M^S) = \frac{1}{n} \sum_{i = 1}^{n}E^k_i(R,M^S).
\end{equation}

Consequently, the repetition error $E^k$ also takes values in the interval $[0, 1]$, where $0$ means perfect execution across all frames and $1$ signifies maximum error throughout the entire repetition.

Our ultimate goal is to decide whether we can reliably separate correct and incorrect repetitions. Let us notice that even correct repetitions can have non-zero feature errors due to imprecise pose detections or slight hesitations of the exercising person. However, if the difference between the error values of the correct and incorrect repetitions is sufficiently large, it is possible to define a separation threshold. To capture this aspect, we introduce a new \emph{feature separability metric} $\Psi^k$ that, for a given feature $F_k$, compares the error of a given repetition $R$ to the average error of correct repetitions $Reps^{S}_{OK}$ that formed the model $M^S$:

\begin{equation}
\label{eq:SeparabilityScore}
  \Psi^k(R,M^S,Reps^{S}_{OK}) = E^k(R,M^S) - \frac{1}{\left|Reps^{S}_{OK}\right|}\sum_{R_i \in Reps^{S}_{OK}}E^k(R_i,M^S)
\end{equation}

To evaluate a specific camera viewpoint at the exercise level, we compute the overall \emph{separability metric} $\Psi$ by averaging the individual scores $\Psi^k(R, M^S, Reps^S_{OK})$ across all erroneous repetitions $R$ and their respective features.

The feature separability metric $\Psi^k$ theoretically ranges from $[-1, 1]$. A value of 1 represents the ideal case, in which the incorrect repetition obtains the maximum possible error ($E^k=1$), while all correct repetitions have zero error ($E^k=0$).
$\Psi = 0$ represents the baseline threshold where bad movements are indistinguishable from normal execution noise, while $\Psi < 0$ indicates inverted error detection (i.e., where correct repetitions are penalized more than erroneous ones). Because any practical system requires $\Psi > 0$ to function, we define the valid operational region for discriminative quality as $\Psi \in (0; 1]$.

\subsection{Problem Formulation and Research Questions}

\begin{figure}[!ht]
    \centering
    \includegraphics[width=\linewidth]{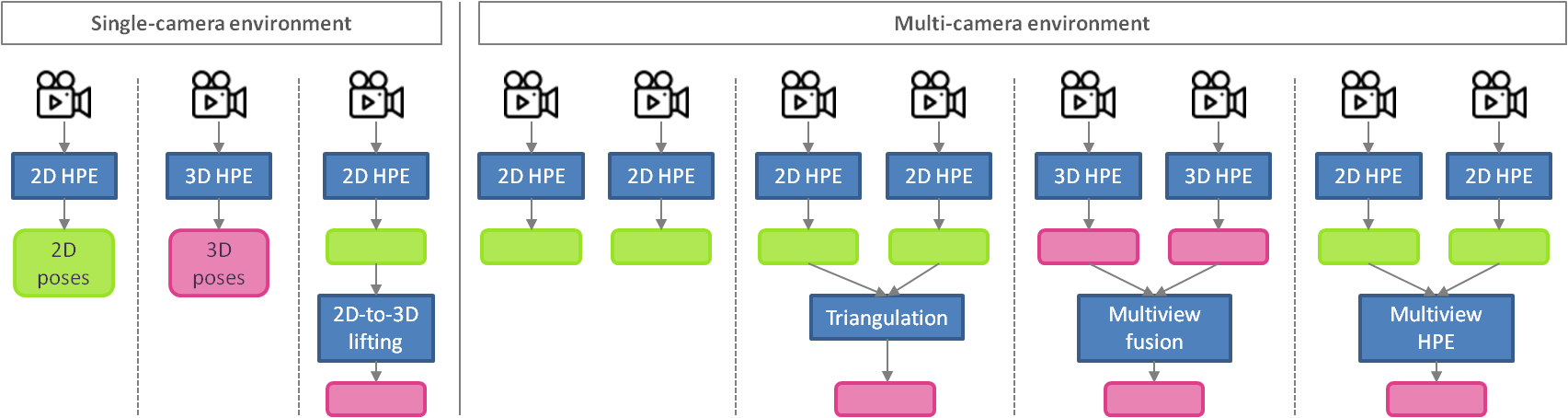}
    \caption{Possible approaches to pose data estimation in single and multiple camera environments.}
    \label{fig:approaches}
\end{figure}

The quality of exercise monitoring is strongly influenced by the accuracy and reliability of the skeletal data extracted from RGB video. For both single- and multi-camera settings, several pose acquisition strategies exist, as illustrated in Fig.~5. In a single-camera environment, the main design choices involve selecting an appropriate camera viewpoint and determining whether 2D or 3D human pose estimation (HPE) provides a more suitable representation for exercise assessment. In a multi-camera environment, we consider a practical setup with two cameras placed approximately orthogonally to each other. However, multiple strategies need be considered for combining the available views, ranging from independent 2D analysis to the reconstruction or estimation of 3D skeletal representations.

The objective of this study is to determine which combinations of camera placement and skeleton acquisition strategies provide the most stable and discriminative information for personalized exercise quality assessment. Specifically, we investigate the following research questions:

\begin{enumerate}[label=\emph{RQ\arabic*:}]
    \item \emph{Feature-level analysis}. How does camera viewpoint influence the accuracy of individual exercise features?
    \item \emph{Exercise-level optimization}. How can the optimal camera viewpoint be determined for a given rehabilitation exercise?
    \item \emph{Orientation robustness}. Given an optimal camera viewpoint, how sensitive is exercise quality assessment to small deviations in patient orientation?
    \item \emph{Pose dimensionality}. Does 2D or 3D skeletal representation provide more reliable information for exercise quality assessment?

    \item \emph{Multi-camera fusion}. What is the most suitable strategy for obtaining 3D skeletal representations when two uncalibrated cameras are available?
    \item \emph{Multi-camera benefit}. What improvement in exercise quality assessment can be achieved by using two cameras instead of one?
\end{enumerate}

RQ1--RQ4 focus on single-camera rehabilitation monitoring and analyze how viewpoint selection and skeleton representation influence both individual movement features and overall exercise assessment. RQ5 and RQ6 address multi-camera scenarios, focusing respectively on combining information from multiple views and quantifying the practical benefit of additional camera input.

The following Sections~4.3 and~4.4 describe the investigated single- and multi-camera skeleton acquisition strategies, including their specific implementations. The research questions are subsequently addressed through experiments conducted on the REHAB26-ViewAngles dataset in Section~5.

\subsection{Single Camera Environments}

When only a single video stream is available, the quality of skeleton data is determined solely by the accuracy of a given human-pose-estimation (HPE) method and the camera angle. As discussed earlier, there are two types of skeleton data that HPE methods can provide: 2D coordinates, expressed in pixels within the video frame, and 3D coordinates, expressed in virtual meters. Depending on the specific HPE architecture, the 3D coordinates can be derived directly from the video, or a two-stage approach can be used, where 2D joint locations are first detected and then lifted to the 3D space.

The suitability of 2D and 3D HPE for different tasks remains an open question. While 3D poses principally enable all-directional assessment of movement quality, extracting them is difficult, with depth as the main source of uncertainty. Using 2D pose estimation avoids this problem, but at the cost of observing the movement only through a single projection, where what can be seen and measured depends strongly on the camera viewpoint.

In practice, patients often struggle to position themselves correctly, and joint occlusions are common. Such occlusions reduce both the reliability of HPE and, subsequently, that of feature extraction, making the assessment of exercise quality difficult, as apparent similarities may arise by chance rather than from accurate execution. Furthermore, different exercises exhibit distinct kinematic characteristics that favor particular viewing angles. Lateral views may better capture squat depth and knee alignment, while frontal views might optimally reveal arm abduction symmetry and shoulder positioning. This paper, therefore, addresses a critical practical deployment question: which single camera position provides the most informative view when only one camera is available, as is typically the case in home rehabilitation settings?

\emph{Implementation.} For single-camera HPE, we utilize the widely adopted MediaPipe Pose model~\footnote{\url{https://developers.google.com/edge/mediapipe/solutions/vision/pose_landmarker}}, which offers both 2D and 3D pose estimates. To ensure that our findings regarding optimal viewing angles are not merely artifacts of a single network architecture, we cross-validate our 2D results using YOLO26 \citep{sapkota2025yolo26}. To evaluate these pose estimates, we define the extracted features for individual coordinate spaces as follows: for 2D data, we compute the feature angles directly in the camera's field of view; for 3D data, we first normalize each pose so that the person's hips are aligned with the $x$-axis, and then measure the feature angles in either the $xy$, $yz$, or $xz$ plane, depending on the specific feature type.

\subsection{Multiple Camera Environments}

As depicted in Fig.~\ref{fig:approaches}, we identified 4 approaches to analyzing human motion recorded by 2 synchronized cameras. In the first method, a 2D skeleton pose is extracted from each stream, and individual features are computed from the more suitable view. The other three options utilize different techniques to create a more precise 3D skeleton pose.

\textbf{Orthogonal 2D HPE.} 2D HPE is known to provide sufficiently precise estimates of joint coordinates in the camera view plane. With two cameras, we can capture orthogonal views of the exercising person so that each exercise feature is clearly visible in at least one view. For each feature, it is determined in advance which view should be considered.

\emph{Implementation.} We utilize the MediaPipe Pose model for all 2D pose extractions in this section.

\textbf{Arithmetic Triangulation.} Geometric triangulation is a standard approach for reconstructing 3D joint positions by intersecting back-projected rays from multiple 2D camera views~\citep{hartley1997triangulation}.
This approach requires explicit camera calibration. Because our recording setup was highly controlled, using identical, time-synchronized cameras placed at exact $45^\circ$ intervals, we derived the necessary calibration parameters directly from the known physical geometry. In practice, minor 2D tracking noise prevents perfect ray intersections, so the final 3D coordinate is estimated by finding the optimal geometric fit.

\emph{Implementation.} We utilize the OpenCV library \citep{bradski2000opencv}, which follows the formulation to compute the Direct Linear Transform (DLT)~\citep{hartley2003multiple}.

\textbf{Weighted Multi-View Fusion.} Another option is to extract a 3D pose estimate from each camera and combine these estimates by weighted averaging. Many HPE methods, including MediaPipe, provide confidence scores for each estimated joint position, which can be used as the weight. The underlying assumption is that joints observed from a more obstructed viewpoint receive lower visibility scores and should contribute less to the final estimate.

\emph{Implementation.} We utilize MediaPipe to obtain two estimates of the 3D skeleton pose. Before fusion, we normalize both 3D poses by shifting the root joint to the origin (0,0,0), rotating the skeletons to a frontal orientation, and scaling them to a common height. This ensures that joint positions from different viewpoints are comparable. The merged 3D position of each joint $J_j$ is then computed as a weighted average across cameras $Cam1$ and $Cam3$:
\begin{equation}
J_j = \frac{\sum_{c=1}^{2} ({(W^{c}_{j})}^2 \cdot J^{c}_{j})}{\sum_{c=1}^{2} {(W^{c}_{j})}^2},
\end{equation}
where $J^{c}_{j}$ is the 3D position of joint $J_j$ from camera $c$, and $W^{c}_{j}$ is its weight (the visibility score). Squared weighting emphasizes joints with higher confidence and reduces the influence of occluded joints.

\textbf{Multi-View HPE.} We adopt the Ray-Based Universal Multi-view Pose Lifter (RUMPL)~\citep{ghasemzadeh2026rumpl} as our learning-based fusion backbone. RUMPL’s ray-based transformer architecture offers a robust alternative to geometric fusion, capable of resolving depth ambiguities and handling severe occlusions even in sparse, two-camera configurations. Its ability to learn human kinematics from synthetic data provides a strong, data-efficient baseline for evaluating clinical movements, particularly when real-world multi-view data are limited.

\emph{Implementation.} To ensure comparability with our monocular pipelines, we adapted RUMPL to consume 2D joint detections from MediaPipe. Because MediaPipe’s native 25-keypoint set does not fully align with the AMASS ground-truth annotations, we restricted the model to predict the 16 joints that have direct correspondences in the existing pipeline, leaving RUMPL’s core ray-fusion transformer unchanged. For training, we followed the original RUMPL pipeline: synthetic AMASS~\citep{DBLP:conf/iccv/MahmoodGTPB19} poses are rendered from randomized camera positions and fed through the same multi-view ray-triangulation procedure to produce training tokens. At inference, MediaPipe detections from our calibrated cameras are projected into this same ray space using the known camera calibration matrices of our experimental setup.

\section{Experimental Evaluation}

The objective of the experimental evaluation is to determine which camera viewpoints and pose representations most reliably capture exercise-specific movement features and allow us to detect incorrectly performed repetitions. Towards this end, we systematically evaluate different viewing angles and pose estimation methods.

\subsection{Methodology}

The basic unit of exercise quality assessment is the exercise sequence $S$. Let us recall that each sequence in the REHAB26-ViewAngles dataset is composed of 5 correctly performed and 5 intentionally erroneous exercise repetitions. The objective of quality assessment is to assign a low error score to correct repetitions and a high error score to incorrect ones. The ability to distinguish between correct and incorrect execution is quantified by the separability $\Psi$ introduced in Section~\ref{sec:Formalisms}. The score is computed for each incorrect repetition and takes values in the interval $[-1,1]$, where values close to 1 indicate good separation between correct and incorrect execution.

For every exercise sequence, a personalized exercise model is first constructed from the five correct repetitions, as described in Section~\ref{sec:Formalisms}. This model is subsequently compared with each incorrect repetition in the sequence. Let us recall that each incorrect repetition corresponds to a predefined clinical error and is associated with one or more features designed to detect that particular error. In experiments that focus on individual features, we therefore evaluate $\Psi^k(R^k_{err},M^S,Reps^{S}_{OK})$, where $R^k_{err}$ is the erroneous repetition with error on feature $F_k$. In experiments evaluating the overall suitability of a camera configuration for a given exercise, all erroneous repetitions within the sequence are considered, and the overall separability $\Psi$ is computed for each repetition across all relevant features. The relevant features were identified using ground-truth annotations specifying the error type for each repetition.

The separability score $\Psi$ is computed independently for every exercise sequence and every considered skeleton acquisition method. Each sequence was recorded simultaneously by three cameras, providing three single-camera viewpoints. In addition, one orthogonal two-camera configuration, comprising Cam1 and~Cam3, is considered for multi-camera experiments. To evaluate a particular viewpoint, the corresponding separability scores are averaged over all exercise sequences recorded from that viewpoint.

For single-camera experiments, all viewing angles available in the REHAB26-ViewAngles dataset are evaluated, spanning the interval from $-90^\circ$ to $90^\circ$. As defined in Section~\ref{sec:AcquisitionProtocol}, $0^\circ$ denotes a frontal view, while positive and negative angles correspond to viewpoints on the participant's right- and left-hand sides, respectively. For the orthogonal two-camera configuration, the viewing angle is specified with respect to Cam1, yielding a range from $-90^\circ$ to $0^\circ$ (see Fig.~\ref{fig:datasetRecording}).

\subsection{Single Camera Environments}

A comprehensive set of experiments over the REHAB26-ViewAngles dataset allows us to address all research questions we set out to investigate. The first three questions for single-camera settings analyze effective camera angles and sensitivity to viewpoint changes using 2D skeleton data, while the forth question discusses the impact of using 3D pose estimates instead of 2D.

\begin{figure}[!ht]
    \centering
    \includegraphics[width=\linewidth]{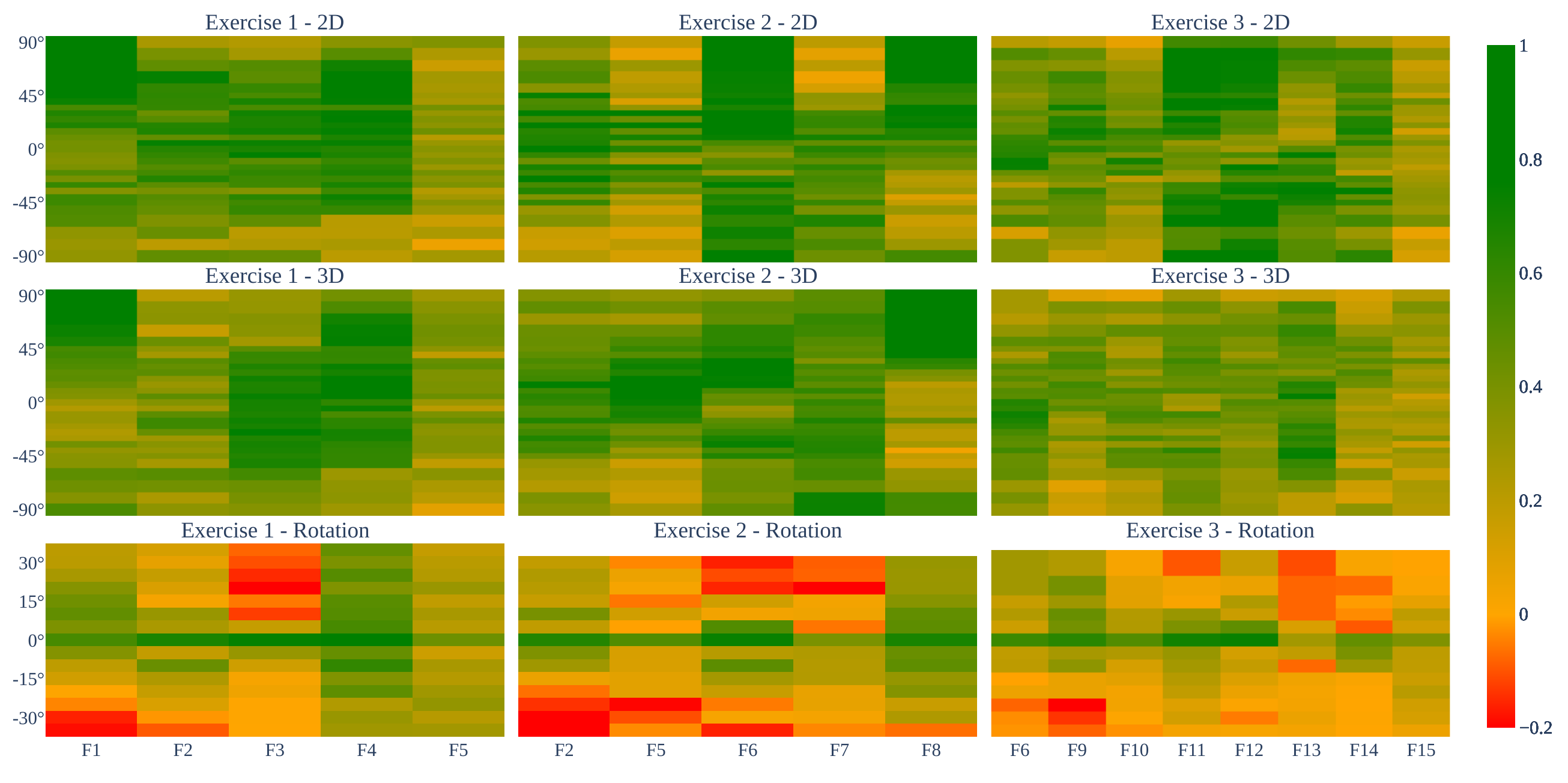}
    \caption{Illustration of \textit{feature separability scores} ($\Psi^k$) for Exercises 1, 2, and 3 at different viewpoint angles. In each heatmap, rows correspond to camera angles and columns to individual features. The uppermost heatmaps present features computed from 2D poses. The second row of heatmaps shows features computed from 3D poses. In the last row, rotation heatmaps visualize how deviations from the optimal viewing angle affect the separability scores. Darker green indicates higher separability scores, while dark red corresponds to situations where correct and incorrect repetitions are indistinguishable.}
    \label{fig:heatmap}
\end{figure}

\textbf{RQ1: Feature-level analysis.}
To understand the effect of camera angles on the visibility of individual features, we utilize \emph{heatmaps} presented in Fig.~\ref{fig:heatmap} (first row), which visualize feature separability scores ($\Psi^k$) across viewing angles. The results clearly demonstrate that the optimal viewpoint is feature-dependent. For example, \textit{shoulder elevation} (F2) is most reliably assessed from a frontal view ($0^\circ$), whereas the \textit{knee pass toes} features (F11, F12) require an oblique or lateral view, with the highest separability typically achieved at angles of $45^\circ$ or greater. Furthermore, for Exercises~1 and~2, positive viewing angles generally outperform negative ones because the movements are performed with the right arm, making the active limb more visible from the participant's right-hand side.

Although a few isolated outliers can be observed (e.g., a performance peak of F3 at $-90^\circ$), these are likely caused by incidental postural changes rather than the camera viewpoint itself. More importantly, the heatmaps show that most features exhibit a relatively broad range of viewing angles with comparable separability scores. Consequently, reliable feature extraction does not require precise patient positioning, an observation that is further investigated in RQ3.

Beyond spatial considerations, the heatmaps also reveal an important temporal aspect of exercise quality assessment. Features such as \textit{insufficient arm abduction} (F5) appear comparatively weak, with lower separability scores across all viewing angles than, for example, the \textit{knee pass toes} features (F11, F12). This, however, does not necessarily indicate that the feature itself is less informative. Rather, the corresponding movement error occurs only briefly near the peak of the repetition, so although the feature correctly captures the error at that instant, its average error over the entire repetition remains relatively low. In contrast, the \textit{knee pass toes} error persists throughout a larger portion of the squat movement, producing a stronger and more stable separation between correct and incorrect executions. This observation suggests that feature separability is influenced not only by the severity of the detected error but also by its temporal persistence. Developing feature aggregation strategies that jointly account for both factors represents an interesting direction for future research.

\textbf{RQ2: Exercise-level optimization.} To identify the most effective camera placement for each exercise, we evaluated the overall separability $\Psi$ for each camera angle (Table~\ref{tab:exercises}). Based on this aggregation, the optimal consensus viewing angles across all monocular models (across both 2D from MPP and YOLO26) are $15^\circ$ for Exercise 1, $20^\circ$ for Exercise 2, and $30^\circ$ for Exercise 3. While the exact optimal angles for YOLO and MPP estimation show minor variations, they remain highly consistent, indicating that the best viewpoint is exercise-specific rather than dependent on the underlying HPE architecture.

\begin{table*}[!t]
\centering
\caption{Overall separability scores across single (S) and multi-camera (M) methods and various viewpoint angles for all three exercises. For multi-camera approaches, the column header corresponds to the Cam1 viewpoint; the Cam3 viewpoint is always orthogonal (i.e., increased by $90^\circ$). Higher values indicate better distinguishability between correct and incorrect movements. Bold values denote the highest separability score within each row. Underlined values highlight the absolute best score within the single-camera and multi-camera configurations, respectively. The results indicate that a single camera is often sufficient for accurate assessment, provided the patient maintains the optimal standing orientation.}
\label{tab:all_exercises_scores}

\textbf{(a) Exercise 1: Arm Abduction} \vspace{4pt}\\
\resizebox{\textwidth}{!}{\input{plots_new/tab1}}

\vspace{0.5cm} %

\textbf{(b) Exercise 2: Shoulder Extension} \vspace{4pt}\\
\resizebox{\textwidth}{!}{\input{plots_new/tab2}}

\vspace{0.5cm}

\textbf{(c) Exercise 3: Squat} \vspace{4pt}\\
\resizebox{\textwidth}{!}{\input{plots_new/tab3}}
\label{tab:exercises}

\end{table*}

As expected, no single ``universal'' best viewpoint exists. The optimal angle is determined by the combination of discriminative features within each exercise. It is not strongly correlated with the primary movement direction: although Exercise 1 is performed mainly in the frontal plane and Exercise 2 in the lateral plane, their optimal viewpoints are nearly identical. Instead, the decisive factor is the direction of the most frequent execution errors. For Exercises 1 and 2, most errors occur in the frontal plane and are better captured from lower viewing angles, whereas Exercise 3 involves substantial errors in both frontal and lateral planes, making a half-profile view more suitable.
Importantly, an optimally placed 2D camera increases the separability by 16.9\,\% relative to the commonly used $0^\circ$ frontal view. Here, relative improvements are reported as percentage gains over the baseline separability score within the positive region, calculated as $\frac{\Psi_{\text{new}} - \Psi_{\text{base}}}{\Psi_{\text{base}}} \times 100\,\%$. This highlights that camera placement is a decisive factor for maximizing discriminative performance in monocular 2D pose-based assessment.

\textbf{RQ3: Orientation robustness.} In this task, we investigate how slight shifts in a patient's orientation affect the exercise quality assessment. It is indeed likely that the patient's position will differ slightly for individual exercise sessions, so we need the assessments to be robust to small angle changes. To examine this, we create a model $M^S$ for each exercise using the correct repetitions recorded at previously identified ``best angles'' ($15^\circ$ for Exercise 1, $20^\circ$ for Exercise 2, and $30^\circ$ for Exercise 3). Then, we use these models to evaluate repetitions recorded at shifted angles. In particular, we analyze incremental deviations ranging from $-35^\circ$ to $+35^\circ$. The increments are $5^\circ$ where possible with the viewpoints available in REHAB26-ViewAngles (see Table~\ref{tab:dataset_angles}), otherwise $10^\circ$. The results are visualized in the third row of Fig.~\ref{fig:heatmap}, where the ``best angle'' is centered at $0$ to show the impact of the shift itself. The data reveal that most features remain reasonably stable within the window of approximately $5^\circ$ to $10^\circ$ from the optimal orientation. Beyond a $15^\circ$ shift, we observe a rapid deterioration in multiple features. These findings provide an important insight for real-world rehabilitation apps: to maintain accuracy, the system
needs to guide the patients into a correct position with a tolerance of about $10^\circ$, which is achievable using state-of-the-art 3D HPE.

\begin{figure}[!ht]
    \centering
    \includegraphics[width=0.65\linewidth]{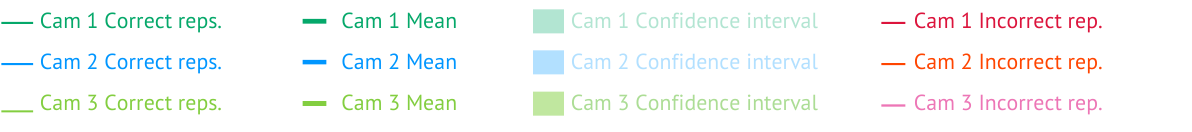}

    \subfloat[Person 1 (2D MediaPipe)]
    {\label{subfig:honza_2d}\includegraphics[width=\linewidth]{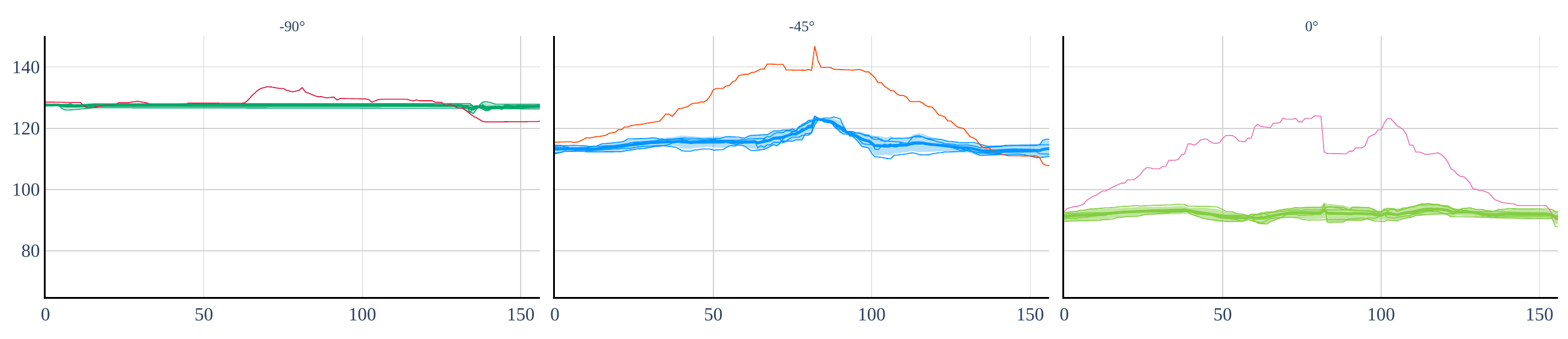}}

    \subfloat[Person 2 (2D MediaPipe)]
    {\label{subfig:petra_2d}\includegraphics[width=\linewidth]{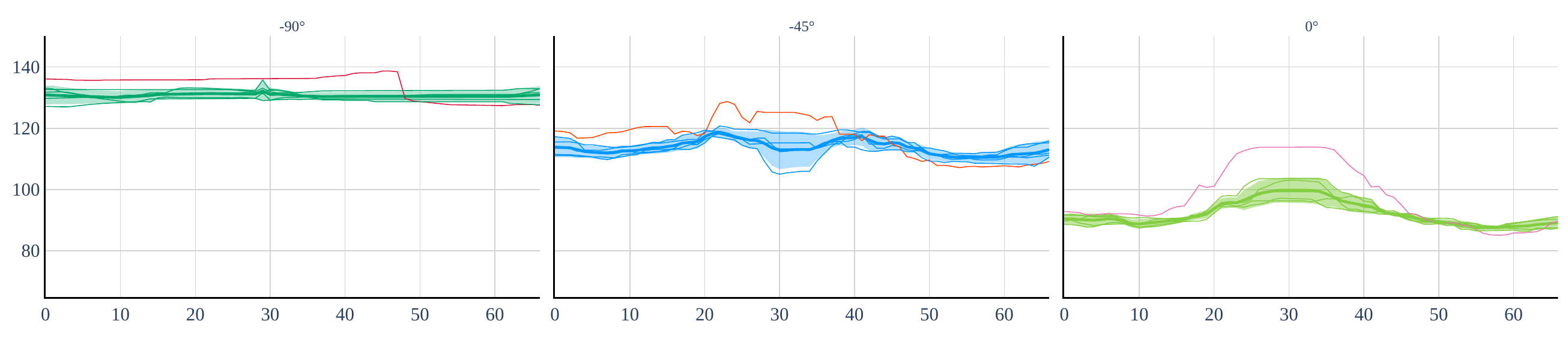}}

    \subfloat[Person 1 (3D MediaPipe)]
    {\label{subfig:honza_3d}\includegraphics[width=\linewidth]{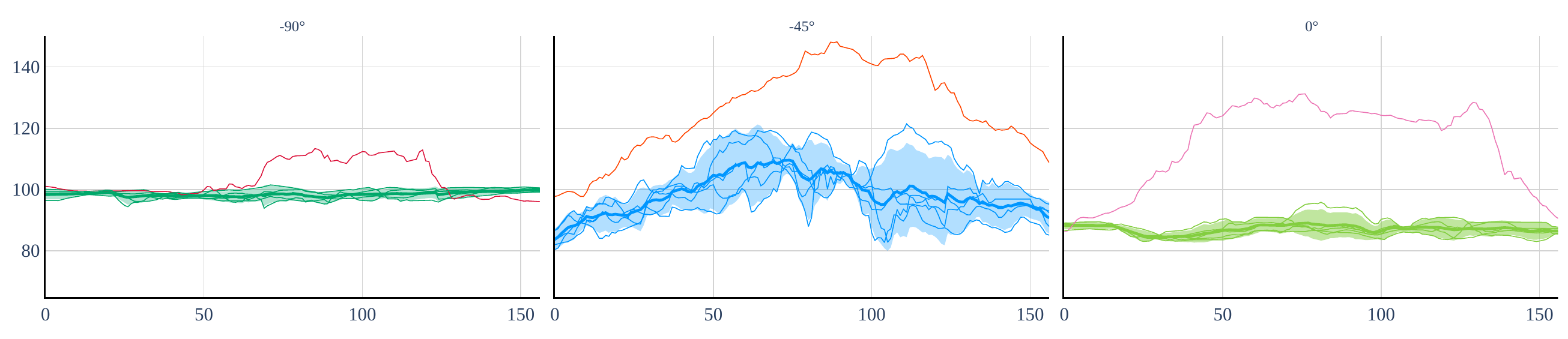}}

   \caption{Repetition analysis of feature F3 (Lateral head tilt) from Exercise 1 across three simultaneous viewpoints ($VP_{Cam1} = -90^{\circ}$, $VP_{Cam2} = -45^{\circ}$, and $VP_{Cam3} = 0^{\circ}$). Baseline values shift distinctly with the camera perspective. In the middle plot of (a), normal repetitions spread out to create a wider confidence interval, yet the erroneous repetition trajectory remains completely separate and easily detectable. Conversely, Person 2 in (b) shows a smaller separation between correct and erroneous trajectories due to a less pronounced error execution, highlighting the need for optimal camera placement. Across all setups, the frontal $0^{\circ}$ viewpoint provides the most distinct signal, while a comparison between (a) and (c) demonstrates that MediaPipe's 3D estimates introduce substantial noise and signal instability compared to 2D tracking.}

    \label{fig:head_tilt_analysis}
\end{figure}

\textbf{RQ4: Pose dimensionality.}
The comparison between 2D and 3D skeletal representations is shown in the second row of Fig.~\ref{fig:heatmap}, with overall separability scores reported in Table~\ref{tab:exercises}. The results indicate only minor differences in stability and discriminative power for Exercises 1 and 2, but a pronounced difference for Exercise 3, where the 2D representations provide clearly better separability. This can likely be attributed to the nature of Exercise 3, which involves larger full-body movements; inaccurate depth estimates in such cases distort the overall pose representation. Overall, the 2D representation yields more consistent and more clearly separable results, aligning with previous findings that suggest greater stability of 2D pose features~\citep{sun2025depth}. For Exercises~1 and~3, the 3D representation performs noticeably worse than its 2D counterpart, indicating that the additional depth dimension introduces pose estimation error that outweighs its benefits. Exercise~2 shows slightly better results in 3D, likely due to its stronger reliance on depth-related motion, though the differences remain subtle. When considered together with the findings from RQ3, the practical usability of 3D skeleton data is further reduced, as changes in camera angle diminish feature differences even more. Overall, despite their intuitive appeal, 3D representations appear less robust than 2D skeleton data for exercise-quality assessment in the evaluated setting. This structural divergence and data degradation are visually highlighted in Fig.~\ref{fig:head_tilt_analysis}, where the 3D extraction signals display significant baseline noise compared to the stable 2D multi-angle tracking.

\subsection{Multiple Camera Environments}
In this section, we address the remaining two research questions that focus on the combination of inputs from two camera streams.

\textbf{RQ5: Multi-camera fusion.} To find out which multi-camera approach can provide the most precise 3D poses, three extraction strategies were evaluated using data from Cam1 and Cam3: 3D triangulation, 3D merging, and RUMPL, a state-of-the-art transformer-based multi-view approach. The results in Table~\ref{tab:exercises} demonstrate that the triangulation-based approach produces the most stable 3D representations and achieves the highest overall improvement among the evaluated methods. By exploiting geometric consistency across views, triangulation reduces depth-related noise and yields more reliable joint estimates.

In contrast, the weighted 3D merging and RUMPL methods show inconsistent performance. The merging strategy remains sensitive to inaccuracies in individual depth estimates, often assigning high confidence to noisy data, while the RUMPL approach, though powerful, is built for data-rich environments and struggles with the depth ambiguity inherent in sparse, two-camera setups. We further observe that performance is highly sensitive to camera placement. The best results occur when both cameras capture the subject from a side-on perspective, at roughly $-40^\circ$ and $50^\circ$. This orientation offers an optimal trade-off, avoiding the self-occlusions typical of frontal or lateral views while maximizing the geometric information available to the system.

\textbf{RQ6: Multi-camera benefit.}
The final research question focuses on the benefits of multi-camera exercise monitoring, as compared to single-camera environments. In Table~\ref{tab:exercises}, we can observe that the only multi-camera approach that consistently meets or outperforms single-camera 2D HPE is the orthogonal combination of two 2D views. The increase of separability over the single camera 2D MPP setup is 0\,\% for Exercise 1, 26.6\,\% for Exercise 2, and 12.7\,\% for Exercise 3. Among the 3D pose estimation methods, the triangulation is the most promising, but it is not reliable for all exercise types -- the respective separability gains are 1.5\,\%, 23.3\,\%, and -1.8\,\%.
Overall, although multi-view approaches can provide moderate performance gains, their practical deployment is more demanding, requiring more space, specialized calibration, and hardware-level camera synchronization that is virtually impossible to achieve with standard smartphones in unconstrained home environments. In contrast, single-view 2D methods are simpler and more flexible, and their fast implementations for mobile devices are already available. Therefore, we recommend single-camera 2D pose estimation as the primary approach for monitoring rehabilitation exercises.

\subsection{Computational Complexity and Real-Time Viability}

To be viable for home rehabilitation, assessment systems must run reliably on consumer smartphones. Standard 2D pose estimation models are well-suited for this constraint: MediaPipe Full requires approximately $10$\,M parameters ($\approx 2$\,GFLOPs), and YOLO26 uses roughly $3.6$\,M parameters ($\approx 10$\,GFLOPs). In contrast, Multi-View Human Pose Estimation (MV-HPE) pipelines like RUMPL did not improve skeleton data quality. While RUMPL's compact 3D lifting architecture ($\approx 12$\,M parameters, $\approx 0.33$\,GFLOPs) could theoretically run on a smartphone, requiring simultaneous 2D feature extraction across multiple video feeds quickly exceeds hardware limits. Among multi-camera setups, the orthogonal 2D approach is the most promising, delivering the highest assessment quality. While running lightweight 2D extraction independently across two smartphones is computationally well within mobile hardware capabilities, the primary deployment challenge remains the temporal synchronization of the video streams to enable joint feature evaluation.

\section{Conclusions}

In this work, we investigate an often overlooked aspect of automatic rehabilitation monitoring: the impact of camera viewpoint on exercise assessment. We introduce the REHAB26-ViewAngles dataset and a dedicated methodology to quantify the separability of exercise errors. Contrary to common practice, which relies on frontal or side views, we find that slight lateral and half-profile perspectives are optimal for three basic exercise types, improving separability by 16.9\,\% compared to the commonly used $0^\circ$ frontal view. We further evaluate robustness under realistic conditions with inconsistent camera placement, observing a tolerance of approximately $10^\circ$; larger deviations substantially reduce assessment accuracy. These findings emphasize the need to guide patients to appropriate camera positions in vision-based rehabilitation systems.

We also analyzed suitable data acquisition strategies for feature extraction in multi-camera environments. Our experiments indicate that 2D pose estimation currently provides more accurate quality assessment than 3D methods, leading to better exercise separability (on average 9.2\,\%). To improve 3D reconstruction, we proposed four multi-camera fusion strategies. Among them, the orthogonal approach achieved the greatest improvement, increasing performance by about 13.1\,\% over the best single-camera setup, while triangulation provided a more modest gain of approximately 7.6\,\%. Furthermore, the orthogonal method is highly advantageous for real-world deployment, as it is computationally lightweight, requires no additional model training, and allows users to simply place two available cameras in the environment to intuitively combine their views. With this approach, the only remaining practical challenge is ensuring robust temporal synchronization between the devices. In contrast, the merged strategy showed greater sensitivity to underlying inaccuracies in 3D pose estimation, while the MV-HPE approach struggled with joint occlusions, as reliably resolving them typically demands denser camera arrays (e.g., four or more viewpoints).

Ultimately, this study provides both methodological and practical contributions to rehabilitation research. Our findings offer actionable guidance for the real-world development of vision-based rehabilitation systems. Future work will extend the feature separability scoring framework to more explicitly account for the influence of error duration. Specifically, we aim to explore dynamic temporal weighting to ensure that errors at the critical apex of a movement are prioritized over minor fluctuations during resting phases.

\section*{Acknowledgment}

This work was supported by the Ministry of Industry and Trade of the Czech Republic under the TWIST programme, project No. FY01010024.

\section*{Declaration of Generative AI and AI-assisted technologies in the writing process}

During the preparation of this work, the author(s) used Claude (Anthropic) and Gemini (Google) in order to improve the grammar, clarity, and flow of the text. After using these tools, the author(s) reviewed and edited the content as needed and take(s) full responsibility for the content of the published article.

\bibliographystyle{plainnat}
\bibliography{references-Jan,references-Jan-disa}

\end{document}

%% file: plots_new/tab1.tex
{\setlength{\tabcolsep}{2.25pt}%
\begin{tabular}{lll|ccccccccccccccc}
Cams & Dims & Method &  -90 & -60 & -40 & -30 & -20 & -15 & -10 & 0 & 10 & 15 & 20 & 30 & 40 & 60 & 90 \\
\hline
\multirow[c]{3}{*}{S} & 2D & MPP & {\cellcolor[HTML]{FECA79}} \color[HTML]{000000} 0.36 & {\cellcolor[HTML]{FEDC88}} \color[HTML]{000000} 0.39 & {\cellcolor[HTML]{EBF7A3}} \color[HTML]{000000} 0.55 & {\cellcolor[HTML]{E3F399}} \color[HTML]{000000} 0.57 & {\cellcolor[HTML]{E6F59D}} \color[HTML]{000000} 0.56 & {\cellcolor[HTML]{E5F49B}} \color[HTML]{000000} 0.57 & {\cellcolor[HTML]{F1F9AC}} \color[HTML]{000000} 0.54 & {\cellcolor[HTML]{E2F397}} \color[HTML]{000000} 0.58 & {\cellcolor[HTML]{DAF08D}} \color[HTML]{000000} 0.60 & {\cellcolor[HTML]{B7E075}} \color[HTML]{000000} \underline{\textbf{0.66}} & {\cellcolor[HTML]{D1EC86}} \color[HTML]{000000} 0.62 & {\cellcolor[HTML]{D1EC86}} \color[HTML]{000000} 0.61 & {\cellcolor[HTML]{D5ED88}} \color[HTML]{000000} 0.61 & {\cellcolor[HTML]{C7E77F}} \color[HTML]{000000} 0.63 & {\cellcolor[HTML]{FEE18D}} \color[HTML]{000000} 0.41 \\
 & 3D & MPP & {\cellcolor[HTML]{FDB365}} \color[HTML]{000000} 0.31 & {\cellcolor[HTML]{FEEDA1}} \color[HTML]{000000} 0.44 & {\cellcolor[HTML]{FFFCBA}} \color[HTML]{000000} 0.49 & {\cellcolor[HTML]{FFF3AC}} \color[HTML]{000000} 0.46 & {\cellcolor[HTML]{FDFEBC}} \color[HTML]{000000} 0.51 & {\cellcolor[HTML]{E2F397}} \color[HTML]{000000} 0.57 & {\cellcolor[HTML]{FEFFBE}} \color[HTML]{000000} 0.50 & {\cellcolor[HTML]{FFF6B0}} \color[HTML]{000000} 0.47 & {\cellcolor[HTML]{F1F9AC}} \color[HTML]{000000} 0.54 & {\cellcolor[HTML]{EEF8A8}} \color[HTML]{000000} 0.55 & {\cellcolor[HTML]{E2F397}} \color[HTML]{000000} 0.57 & {\cellcolor[HTML]{DCF08F}} \color[HTML]{000000} \textbf{0.59} & {\cellcolor[HTML]{FFF7B2}} \color[HTML]{000000} 0.47 & {\cellcolor[HTML]{FDFEBC}} \color[HTML]{000000} 0.51 & {\cellcolor[HTML]{FEE695}} \color[HTML]{000000} 0.42 \\
 & 2D & YOLO & {\cellcolor[HTML]{F57547}} \color[HTML]{F1F1F1} 0.21 & {\cellcolor[HTML]{FED481}} \color[HTML]{000000} 0.38 & {\cellcolor[HTML]{ECF7A6}} \color[HTML]{000000} 0.55 & {\cellcolor[HTML]{DAF08D}} \color[HTML]{000000} 0.60 & {\cellcolor[HTML]{F2FAAE}} \color[HTML]{000000} 0.53 & {\cellcolor[HTML]{E3F399}} \color[HTML]{000000} 0.57 & {\cellcolor[HTML]{F4FAB0}} \color[HTML]{000000} 0.53 & {\cellcolor[HTML]{E6F59D}} \color[HTML]{000000} 0.56 & {\cellcolor[HTML]{D3EC87}} \color[HTML]{000000} 0.61 & {\cellcolor[HTML]{C9E881}} \color[HTML]{000000} 0.63 & {\cellcolor[HTML]{C7E77F}} \color[HTML]{000000} \textbf{0.64} & {\cellcolor[HTML]{CFEB85}} \color[HTML]{000000} 0.62 & {\cellcolor[HTML]{D9EF8B}} \color[HTML]{000000} 0.60 & {\cellcolor[HTML]{DCF08F}} \color[HTML]{000000} 0.59 & {\cellcolor[HTML]{FEE28F}} \color[HTML]{000000} 0.41 \\
\hline
\multirow[c]{4}{*}{M} & 2D & Ortho. & {\cellcolor[HTML]{DFF293}} \color[HTML]{000000} 0.58 & {\cellcolor[HTML]{D1EC86}} \color[HTML]{000000} 0.61 & {\cellcolor[HTML]{BFE47A}} \color[HTML]{000000} 0.65 & {\cellcolor[HTML]{BBE278}} \color[HTML]{000000} \textbf{0.66} & {\cellcolor[HTML]{C7E77F}} \color[HTML]{000000} 0.64 & {\cellcolor[HTML]{000000}} \color[HTML]{F1F1F1} {\cellcolor{white}} \color{black} -- & {\cellcolor[HTML]{CBE982}} \color[HTML]{000000} 0.63 & {\cellcolor[HTML]{B9E176}} \color[HTML]{000000} \textbf{0.66} & {\cellcolor[HTML]{000000}} \color[HTML]{F1F1F1} {\cellcolor{white}} \color{black} -- & {\cellcolor[HTML]{000000}} \color[HTML]{F1F1F1} {\cellcolor{white}} \color{black} -- & {\cellcolor[HTML]{000000}} \color[HTML]{F1F1F1} {\cellcolor{white}} \color{black} -- & {\cellcolor[HTML]{000000}} \color[HTML]{F1F1F1} {\cellcolor{white}} \color{black} -- & {\cellcolor[HTML]{000000}} \color[HTML]{F1F1F1} {\cellcolor{white}} \color{black} -- & {\cellcolor[HTML]{000000}} \color[HTML]{F1F1F1} {\cellcolor{white}} \color{black} -- & {\cellcolor[HTML]{000000}} \color[HTML]{F1F1F1} {\cellcolor{white}} \color{black} -- \\
 & 3D & Triang. & {\cellcolor[HTML]{FEE999}} \color[HTML]{000000} 0.43 & {\cellcolor[HTML]{D3EC87}} \color[HTML]{000000} 0.61 & {\cellcolor[HTML]{B3DF72}} \color[HTML]{000000} \underline{\textbf{0.67}} & {\cellcolor[HTML]{C9E881}} \color[HTML]{000000} 0.63 & {\cellcolor[HTML]{D9EF8B}} \color[HTML]{000000} 0.60 & {\cellcolor[HTML]{000000}} \color[HTML]{F1F1F1} {\cellcolor{white}} \color{black} -- & {\cellcolor[HTML]{D9EF8B}} \color[HTML]{000000} 0.60 & {\cellcolor[HTML]{F1F9AC}} \color[HTML]{000000} 0.54 & {\cellcolor[HTML]{000000}} \color[HTML]{F1F1F1} {\cellcolor{white}} \color{black} -- & {\cellcolor[HTML]{000000}} \color[HTML]{F1F1F1} {\cellcolor{white}} \color{black} -- & {\cellcolor[HTML]{000000}} \color[HTML]{F1F1F1} {\cellcolor{white}} \color{black} -- & {\cellcolor[HTML]{000000}} \color[HTML]{F1F1F1} {\cellcolor{white}} \color{black} -- & {\cellcolor[HTML]{000000}} \color[HTML]{F1F1F1} {\cellcolor{white}} \color{black} -- & {\cellcolor[HTML]{000000}} \color[HTML]{F1F1F1} {\cellcolor{white}} \color{black} -- & {\cellcolor[HTML]{000000}} \color[HTML]{F1F1F1} {\cellcolor{white}} \color{black} -- \\
 & 3D & Merged & {\cellcolor[HTML]{EBF7A3}} \color[HTML]{000000} 0.55 & {\cellcolor[HTML]{C9E881}} \color[HTML]{000000} \textbf{0.63} & {\cellcolor[HTML]{C7E77F}} \color[HTML]{000000} \textbf{0.63} & {\cellcolor[HTML]{D5ED88}} \color[HTML]{000000} 0.61 & {\cellcolor[HTML]{E0F295}} \color[HTML]{000000} 0.58 & {\cellcolor[HTML]{000000}} \color[HTML]{F1F1F1} {\cellcolor{white}} \color{black} -- & {\cellcolor[HTML]{CFEB85}} \color[HTML]{000000} 0.62 & {\cellcolor[HTML]{FAFDB8}} \color[HTML]{000000} 0.51 & {\cellcolor[HTML]{000000}} \color[HTML]{F1F1F1} {\cellcolor{white}} \color{black} -- & {\cellcolor[HTML]{000000}} \color[HTML]{F1F1F1} {\cellcolor{white}} \color{black} -- & {\cellcolor[HTML]{000000}} \color[HTML]{F1F1F1} {\cellcolor{white}} \color{black} -- & {\cellcolor[HTML]{000000}} \color[HTML]{F1F1F1} {\cellcolor{white}} \color{black} -- & {\cellcolor[HTML]{000000}} \color[HTML]{F1F1F1} {\cellcolor{white}} \color{black} -- & {\cellcolor[HTML]{000000}} \color[HTML]{F1F1F1} {\cellcolor{white}} \color{black} -- & {\cellcolor[HTML]{000000}} \color[HTML]{F1F1F1} {\cellcolor{white}} \color{black} -- \\
 & 3D & RUMPL & {\cellcolor[HTML]{FFF2AA}} \color[HTML]{000000} \textbf{0.46} & {\cellcolor[HTML]{FEE797}} \color[HTML]{000000} 0.42 & {\cellcolor[HTML]{FFF3AC}} \color[HTML]{000000} \textbf{0.46} & {\cellcolor[HTML]{FEDC88}} \color[HTML]{000000} 0.39 & {\cellcolor[HTML]{FED884}} \color[HTML]{000000} 0.39 & {\cellcolor[HTML]{000000}} \color[HTML]{F1F1F1} {\cellcolor{white}} \color{black} -- & {\cellcolor[HTML]{FEDC88}} \color[HTML]{000000} 0.39 & {\cellcolor[HTML]{FEDA86}} \color[HTML]{000000} 0.39 & {\cellcolor[HTML]{000000}} \color[HTML]{F1F1F1} {\cellcolor{white}} \color{black} -- & {\cellcolor[HTML]{000000}} \color[HTML]{F1F1F1} {\cellcolor{white}} \color{black} -- & {\cellcolor[HTML]{000000}} \color[HTML]{F1F1F1} {\cellcolor{white}} \color{black} -- & {\cellcolor[HTML]{000000}} \color[HTML]{F1F1F1} {\cellcolor{white}} \color{black} -- & {\cellcolor[HTML]{000000}} \color[HTML]{F1F1F1} {\cellcolor{white}} \color{black} -- & {\cellcolor[HTML]{000000}} \color[HTML]{F1F1F1} {\cellcolor{white}} \color{black} -- & {\cellcolor[HTML]{000000}} \color[HTML]{F1F1F1} {\cellcolor{white}} \color{black} -- \\
\end{tabular}
}

%% file: plots_new/tab2.tex
{\setlength{\tabcolsep}{2.25pt}%
\begin{tabular}{lll|ccccccccccccccc}
Cams & Dims & Method &  -90 & -60 & -40 & -30 & -20 & -15 & -10 & 0 & 10 & 15 & 20 & 30 & 40 & 60 & 90 \\
\hline
\multirow[c]{3}{*}{S} & 2D & MPP & {\cellcolor[HTML]{FBFDBA}} \color[HTML]{000000} 0.51 & {\cellcolor[HTML]{FEE695}} \color[HTML]{000000} 0.42 & {\cellcolor[HTML]{FEEDA1}} \color[HTML]{000000} 0.44 & {\cellcolor[HTML]{E3F399}} \color[HTML]{000000} 0.57 & {\cellcolor[HTML]{DFF293}} \color[HTML]{000000} 0.58 & {\cellcolor[HTML]{E8F59F}} \color[HTML]{000000} 0.56 & {\cellcolor[HTML]{FEEDA1}} \color[HTML]{000000} 0.44 & {\cellcolor[HTML]{EEF8A8}} \color[HTML]{000000} 0.55 & {\cellcolor[HTML]{D1EC86}} \color[HTML]{000000} \textbf{0.61} & {\cellcolor[HTML]{DAF08D}} \color[HTML]{000000} 0.59 & {\cellcolor[HTML]{D7EE8A}} \color[HTML]{000000} 0.60 & {\cellcolor[HTML]{E9F6A1}} \color[HTML]{000000} 0.56 & {\cellcolor[HTML]{D7EE8A}} \color[HTML]{000000} 0.60 & {\cellcolor[HTML]{F7FCB4}} \color[HTML]{000000} 0.52 & {\cellcolor[HTML]{FEE593}} \color[HTML]{000000} 0.42 \\
 & 3D & MPP & {\cellcolor[HTML]{FFFCBA}} \color[HTML]{000000} 0.49 & {\cellcolor[HTML]{FECE7C}} \color[HTML]{000000} 0.36 & {\cellcolor[HTML]{FEE18D}} \color[HTML]{000000} 0.40 & {\cellcolor[HTML]{EBF7A3}} \color[HTML]{000000} 0.55 & {\cellcolor[HTML]{FFF2AA}} \color[HTML]{000000} 0.46 & {\cellcolor[HTML]{D9EF8B}} \color[HTML]{000000} 0.60 & {\cellcolor[HTML]{FFF0A6}} \color[HTML]{000000} 0.45 & {\cellcolor[HTML]{F5FBB2}} \color[HTML]{000000} 0.52 & {\cellcolor[HTML]{EBF7A3}} \color[HTML]{000000} 0.55 & {\cellcolor[HTML]{D5ED88}} \color[HTML]{000000} 0.61 & {\cellcolor[HTML]{D1EC86}} \color[HTML]{000000} 0.61 & {\cellcolor[HTML]{BFE47A}} \color[HTML]{000000} \underline{\textbf{0.65}} & {\cellcolor[HTML]{D9EF8B}} \color[HTML]{000000} 0.60 & {\cellcolor[HTML]{DDF191}} \color[HTML]{000000} 0.59 & {\cellcolor[HTML]{FFF6B0}} \color[HTML]{000000} 0.47 \\
 & 2D & YOLO & {\cellcolor[HTML]{FEEB9D}} \color[HTML]{000000} 0.43 & {\cellcolor[HTML]{FDAF62}} \color[HTML]{000000} 0.30 & {\cellcolor[HTML]{FEE18D}} \color[HTML]{000000} 0.40 & {\cellcolor[HTML]{FDFEBC}} \color[HTML]{000000} 0.50 & {\cellcolor[HTML]{E0F295}} \color[HTML]{000000} 0.58 & {\cellcolor[HTML]{E9F6A1}} \color[HTML]{000000} 0.56 & {\cellcolor[HTML]{FFF1A8}} \color[HTML]{000000} 0.45 & {\cellcolor[HTML]{FEEC9F}} \color[HTML]{000000} 0.44 & {\cellcolor[HTML]{FAFDB8}} \color[HTML]{000000} 0.51 & {\cellcolor[HTML]{D3EC87}} \color[HTML]{000000} \textbf{0.61} & {\cellcolor[HTML]{D3EC87}} \color[HTML]{000000} \textbf{0.61} & {\cellcolor[HTML]{F7FCB4}} \color[HTML]{000000} 0.52 & {\cellcolor[HTML]{DAF08D}} \color[HTML]{000000} 0.60 & {\cellcolor[HTML]{E9F6A1}} \color[HTML]{000000} 0.56 & {\cellcolor[HTML]{FFF8B4}} \color[HTML]{000000} 0.48 \\
\hline
\multirow[c]{4}{*}{M} & 2D & Ortho. & {\cellcolor[HTML]{BFE47A}} \color[HTML]{000000} 0.65 & {\cellcolor[HTML]{E0F295}} \color[HTML]{000000} 0.58 & {\cellcolor[HTML]{CDEA83}} \color[HTML]{000000} 0.62 & {\cellcolor[HTML]{A5D86A}} \color[HTML]{000000} 0.70 & {\cellcolor[HTML]{7FC866}} \color[HTML]{000000} \underline{\textbf{0.76}} & {\cellcolor[HTML]{000000}} \color[HTML]{F1F1F1} {\cellcolor{white}} \color{black} -- & {\cellcolor[HTML]{D5ED88}} \color[HTML]{000000} 0.61 & {\cellcolor[HTML]{C1E57B}} \color[HTML]{000000} 0.65 & {\cellcolor[HTML]{000000}} \color[HTML]{F1F1F1} {\cellcolor{white}} \color{black} -- & {\cellcolor[HTML]{000000}} \color[HTML]{F1F1F1} {\cellcolor{white}} \color{black} -- & {\cellcolor[HTML]{000000}} \color[HTML]{F1F1F1} {\cellcolor{white}} \color{black} -- & {\cellcolor[HTML]{000000}} \color[HTML]{F1F1F1} {\cellcolor{white}} \color{black} -- & {\cellcolor[HTML]{000000}} \color[HTML]{F1F1F1} {\cellcolor{white}} \color{black} -- & {\cellcolor[HTML]{000000}} \color[HTML]{F1F1F1} {\cellcolor{white}} \color{black} -- & {\cellcolor[HTML]{000000}} \color[HTML]{F1F1F1} {\cellcolor{white}} \color{black} -- \\
 & 3D & Triang. & {\cellcolor[HTML]{E3F399}} \color[HTML]{000000} 0.57 & {\cellcolor[HTML]{B1DE71}} \color[HTML]{000000} 0.68 & {\cellcolor[HTML]{8ECF67}} \color[HTML]{000000} \textbf{0.74} & {\cellcolor[HTML]{B5DF74}} \color[HTML]{000000} 0.67 & {\cellcolor[HTML]{B3DF72}} \color[HTML]{000000} 0.68 & {\cellcolor[HTML]{000000}} \color[HTML]{F1F1F1} {\cellcolor{white}} \color{black} -- & {\cellcolor[HTML]{AFDD70}} \color[HTML]{000000} 0.68 & {\cellcolor[HTML]{C9E881}} \color[HTML]{000000} 0.63 & {\cellcolor[HTML]{000000}} \color[HTML]{F1F1F1} {\cellcolor{white}} \color{black} -- & {\cellcolor[HTML]{000000}} \color[HTML]{F1F1F1} {\cellcolor{white}} \color{black} -- & {\cellcolor[HTML]{000000}} \color[HTML]{F1F1F1} {\cellcolor{white}} \color{black} -- & {\cellcolor[HTML]{000000}} \color[HTML]{F1F1F1} {\cellcolor{white}} \color{black} -- & {\cellcolor[HTML]{000000}} \color[HTML]{F1F1F1} {\cellcolor{white}} \color{black} -- & {\cellcolor[HTML]{000000}} \color[HTML]{F1F1F1} {\cellcolor{white}} \color{black} -- & {\cellcolor[HTML]{000000}} \color[HTML]{F1F1F1} {\cellcolor{white}} \color{black} -- \\
 & 3D & Merged & {\cellcolor[HTML]{D9EF8B}} \color[HTML]{000000} 0.60 & {\cellcolor[HTML]{FBFDBA}} \color[HTML]{000000} 0.51 & {\cellcolor[HTML]{DDF191}} \color[HTML]{000000} 0.59 & {\cellcolor[HTML]{C7E77F}} \color[HTML]{000000} \textbf{0.64} & {\cellcolor[HTML]{FFFBB8}} \color[HTML]{000000} 0.49 & {\cellcolor[HTML]{000000}} \color[HTML]{F1F1F1} {\cellcolor{white}} \color{black} -- & {\cellcolor[HTML]{E9F6A1}} \color[HTML]{000000} 0.56 & {\cellcolor[HTML]{E5F49B}} \color[HTML]{000000} 0.57 & {\cellcolor[HTML]{000000}} \color[HTML]{F1F1F1} {\cellcolor{white}} \color{black} -- & {\cellcolor[HTML]{000000}} \color[HTML]{F1F1F1} {\cellcolor{white}} \color{black} -- & {\cellcolor[HTML]{000000}} \color[HTML]{F1F1F1} {\cellcolor{white}} \color{black} -- & {\cellcolor[HTML]{000000}} \color[HTML]{F1F1F1} {\cellcolor{white}} \color{black} -- & {\cellcolor[HTML]{000000}} \color[HTML]{F1F1F1} {\cellcolor{white}} \color{black} -- & {\cellcolor[HTML]{000000}} \color[HTML]{F1F1F1} {\cellcolor{white}} \color{black} -- & {\cellcolor[HTML]{000000}} \color[HTML]{F1F1F1} {\cellcolor{white}} \color{black} -- \\
 & 3D & RUMPL & {\cellcolor[HTML]{FED481}} \color[HTML]{000000} 0.38 & {\cellcolor[HTML]{FEEDA1}} \color[HTML]{000000} 0.45 & {\cellcolor[HTML]{F7FCB4}} \color[HTML]{000000} \textbf{0.52} & {\cellcolor[HTML]{FFF0A6}} \color[HTML]{000000} 0.45 & {\cellcolor[HTML]{FFF3AC}} \color[HTML]{000000} 0.46 & {\cellcolor[HTML]{000000}} \color[HTML]{F1F1F1} {\cellcolor{white}} \color{black} -- & {\cellcolor[HTML]{FED07E}} \color[HTML]{000000} 0.37 & {\cellcolor[HTML]{FDC574}} \color[HTML]{000000} 0.34 & {\cellcolor[HTML]{000000}} \color[HTML]{F1F1F1} {\cellcolor{white}} \color{black} -- & {\cellcolor[HTML]{000000}} \color[HTML]{F1F1F1} {\cellcolor{white}} \color{black} -- & {\cellcolor[HTML]{000000}} \color[HTML]{F1F1F1} {\cellcolor{white}} \color{black} -- & {\cellcolor[HTML]{000000}} \color[HTML]{F1F1F1} {\cellcolor{white}} \color{black} -- & {\cellcolor[HTML]{000000}} \color[HTML]{F1F1F1} {\cellcolor{white}} \color{black} -- & {\cellcolor[HTML]{000000}} \color[HTML]{F1F1F1} {\cellcolor{white}} \color{black} -- & {\cellcolor[HTML]{000000}} \color[HTML]{F1F1F1} {\cellcolor{white}} \color{black} -- \\
\end{tabular}
}

%% file: plots_new/tab3.tex
{\setlength{\tabcolsep}{2.25pt}%
\begin{tabular}{lll|ccccccccccccccc}
Cams & Dims & Method &  -90 & -60 & -40 & -30 & -20 & -15 & -10 & 0 & 10 & 15 & 20 & 30 & 40 & 60 & 90 \\
\hline
\multirow[c]{3}{*}{S} & 2D & MPP & {\cellcolor[HTML]{F5FBB2}} \color[HTML]{000000} 0.53 & {\cellcolor[HTML]{E8F59F}} \color[HTML]{000000} \textbf{0.56} & {\cellcolor[HTML]{ECF7A6}} \color[HTML]{000000} 0.55 & {\cellcolor[HTML]{FDFEBC}} \color[HTML]{000000} 0.51 & {\cellcolor[HTML]{FEDE89}} \color[HTML]{000000} 0.40 & {\cellcolor[HTML]{FFF8B4}} \color[HTML]{000000} 0.48 & {\cellcolor[HTML]{FDBD6D}} \color[HTML]{000000} 0.33 & {\cellcolor[HTML]{FEE999}} \color[HTML]{000000} 0.43 & {\cellcolor[HTML]{FAFDB8}} \color[HTML]{000000} 0.51 & {\cellcolor[HTML]{F7FCB4}} \color[HTML]{000000} 0.52 & {\cellcolor[HTML]{FAFDB8}} \color[HTML]{000000} 0.52 & {\cellcolor[HTML]{E9F6A1}} \color[HTML]{000000} 0.55 & {\cellcolor[HTML]{F1F9AC}} \color[HTML]{000000} 0.54 & {\cellcolor[HTML]{EFF8AA}} \color[HTML]{000000} 0.54 & {\cellcolor[HTML]{FDFEBC}} \color[HTML]{000000} 0.51 \\
 & 3D & MPP & {\cellcolor[HTML]{FB9D59}} \color[HTML]{000000} 0.28 & {\cellcolor[HTML]{FECE7C}} \color[HTML]{000000} 0.36 & {\cellcolor[HTML]{FFF5AE}} \color[HTML]{000000} 0.47 & {\cellcolor[HTML]{FFF1A8}} \color[HTML]{000000} 0.45 & {\cellcolor[HTML]{FEE593}} \color[HTML]{000000} 0.42 & {\cellcolor[HTML]{FFFAB6}} \color[HTML]{000000} 0.48 & {\cellcolor[HTML]{FFFBB8}} \color[HTML]{000000} \textbf{0.49} & {\cellcolor[HTML]{FEEFA3}} \color[HTML]{000000} 0.45 & {\cellcolor[HTML]{FFF8B4}} \color[HTML]{000000} 0.48 & {\cellcolor[HTML]{FEEFA3}} \color[HTML]{000000} 0.45 & {\cellcolor[HTML]{FFF2AA}} \color[HTML]{000000} 0.46 & {\cellcolor[HTML]{FFF8B4}} \color[HTML]{000000} 0.48 & {\cellcolor[HTML]{FED07E}} \color[HTML]{000000} 0.37 & {\cellcolor[HTML]{FEE695}} \color[HTML]{000000} 0.42 & {\cellcolor[HTML]{F57547}} \color[HTML]{F1F1F1} 0.21 \\
 & 2D & YOLO & {\cellcolor[HTML]{F5FBB2}} \color[HTML]{000000} 0.53 & {\cellcolor[HTML]{E5F49B}} \color[HTML]{000000} 0.57 & {\cellcolor[HTML]{EEF8A8}} \color[HTML]{000000} 0.55 & {\cellcolor[HTML]{F8FCB6}} \color[HTML]{000000} 0.52 & {\cellcolor[HTML]{FEE08B}} \color[HTML]{000000} 0.40 & {\cellcolor[HTML]{FAFDB8}} \color[HTML]{000000} 0.51 & {\cellcolor[HTML]{FECA79}} \color[HTML]{000000} 0.36 & {\cellcolor[HTML]{FED683}} \color[HTML]{000000} 0.38 & {\cellcolor[HTML]{E8F59F}} \color[HTML]{000000} 0.56 & {\cellcolor[HTML]{E9F6A1}} \color[HTML]{000000} 0.56 & {\cellcolor[HTML]{ECF7A6}} \color[HTML]{000000} 0.55 & {\cellcolor[HTML]{D7EE8A}} \color[HTML]{000000} \underline{\textbf{0.60}} & {\cellcolor[HTML]{FAFDB8}} \color[HTML]{000000} 0.51 & {\cellcolor[HTML]{EEF8A8}} \color[HTML]{000000} 0.55 & {\cellcolor[HTML]{FFFEBE}} \color[HTML]{000000} 0.50 \\
\hline
\multirow[c]{4}{*}{M} & 2D & Ortho. & {\cellcolor[HTML]{D3EC87}} \color[HTML]{000000} 0.61 & {\cellcolor[HTML]{D3EC87}} \color[HTML]{000000} 0.61 & {\cellcolor[HTML]{D1EC86}} \color[HTML]{000000} \underline{\textbf{0.62}} & {\cellcolor[HTML]{E0F295}} \color[HTML]{000000} 0.58 & {\cellcolor[HTML]{F4FAB0}} \color[HTML]{000000} 0.53 & {\cellcolor[HTML]{000000}} \color[HTML]{F1F1F1} {\cellcolor{white}} \color{black} -- & {\cellcolor[HTML]{E0F295}} \color[HTML]{000000} 0.58 & {\cellcolor[HTML]{D3EC87}} \color[HTML]{000000} 0.61 & {\cellcolor[HTML]{000000}} \color[HTML]{F1F1F1} {\cellcolor{white}} \color{black} -- & {\cellcolor[HTML]{000000}} \color[HTML]{F1F1F1} {\cellcolor{white}} \color{black} -- & {\cellcolor[HTML]{000000}} \color[HTML]{F1F1F1} {\cellcolor{white}} \color{black} -- & {\cellcolor[HTML]{000000}} \color[HTML]{F1F1F1} {\cellcolor{white}} \color{black} -- & {\cellcolor[HTML]{000000}} \color[HTML]{F1F1F1} {\cellcolor{white}} \color{black} -- & {\cellcolor[HTML]{000000}} \color[HTML]{F1F1F1} {\cellcolor{white}} \color{black} -- & {\cellcolor[HTML]{000000}} \color[HTML]{F1F1F1} {\cellcolor{white}} \color{black} -- \\
 & 3D & Triang. & {\cellcolor[HTML]{EFF8AA}} \color[HTML]{000000} \textbf{0.54} & {\cellcolor[HTML]{FFF2AA}} \color[HTML]{000000} 0.46 & {\cellcolor[HTML]{FED481}} \color[HTML]{000000} 0.38 & {\cellcolor[HTML]{FEEC9F}} \color[HTML]{000000} 0.44 & {\cellcolor[HTML]{FFF0A6}} \color[HTML]{000000} 0.45 & {\cellcolor[HTML]{000000}} \color[HTML]{F1F1F1} {\cellcolor{white}} \color{black} -- & {\cellcolor[HTML]{F5FBB2}} \color[HTML]{000000} 0.53 & {\cellcolor[HTML]{FEEC9F}} \color[HTML]{000000} 0.44 & {\cellcolor[HTML]{000000}} \color[HTML]{F1F1F1} {\cellcolor{white}} \color{black} -- & {\cellcolor[HTML]{000000}} \color[HTML]{F1F1F1} {\cellcolor{white}} \color{black} -- & {\cellcolor[HTML]{000000}} \color[HTML]{F1F1F1} {\cellcolor{white}} \color{black} -- & {\cellcolor[HTML]{000000}} \color[HTML]{F1F1F1} {\cellcolor{white}} \color{black} -- & {\cellcolor[HTML]{000000}} \color[HTML]{F1F1F1} {\cellcolor{white}} \color{black} -- & {\cellcolor[HTML]{000000}} \color[HTML]{F1F1F1} {\cellcolor{white}} \color{black} -- & {\cellcolor[HTML]{000000}} \color[HTML]{F1F1F1} {\cellcolor{white}} \color{black} -- \\
 & 3D & Merged & {\cellcolor[HTML]{FFF0A6}} \color[HTML]{000000} 0.45 & {\cellcolor[HTML]{F8FCB6}} \color[HTML]{000000} \textbf{0.52} & {\cellcolor[HTML]{FFFBB8}} \color[HTML]{000000} 0.49 & {\cellcolor[HTML]{FFFAB6}} \color[HTML]{000000} 0.48 & {\cellcolor[HTML]{FFF0A6}} \color[HTML]{000000} 0.45 & {\cellcolor[HTML]{000000}} \color[HTML]{F1F1F1} {\cellcolor{white}} \color{black} -- & {\cellcolor[HTML]{FFFAB6}} \color[HTML]{000000} 0.48 & {\cellcolor[HTML]{FDB768}} \color[HTML]{000000} 0.32 & {\cellcolor[HTML]{000000}} \color[HTML]{F1F1F1} {\cellcolor{white}} \color{black} -- & {\cellcolor[HTML]{000000}} \color[HTML]{F1F1F1} {\cellcolor{white}} \color{black} -- & {\cellcolor[HTML]{000000}} \color[HTML]{F1F1F1} {\cellcolor{white}} \color{black} -- & {\cellcolor[HTML]{000000}} \color[HTML]{F1F1F1} {\cellcolor{white}} \color{black} -- & {\cellcolor[HTML]{000000}} \color[HTML]{F1F1F1} {\cellcolor{white}} \color{black} -- & {\cellcolor[HTML]{000000}} \color[HTML]{F1F1F1} {\cellcolor{white}} \color{black} -- & {\cellcolor[HTML]{000000}} \color[HTML]{F1F1F1} {\cellcolor{white}} \color{black} -- \\
 & 3D & RUMPL & {\cellcolor[HTML]{F8FCB6}} \color[HTML]{000000} \textbf{0.52} & {\cellcolor[HTML]{FED884}} \color[HTML]{000000} 0.38 & {\cellcolor[HTML]{FFF0A6}} \color[HTML]{000000} 0.45 & {\cellcolor[HTML]{FEEB9D}} \color[HTML]{000000} 0.44 & {\cellcolor[HTML]{FEE28F}} \color[HTML]{000000} 0.41 & {\cellcolor[HTML]{000000}} \color[HTML]{F1F1F1} {\cellcolor{white}} \color{black} -- & {\cellcolor[HTML]{FEE18D}} \color[HTML]{000000} 0.40 & {\cellcolor[HTML]{FDAF62}} \color[HTML]{000000} 0.30 & {\cellcolor[HTML]{000000}} \color[HTML]{F1F1F1} {\cellcolor{white}} \color{black} -- & {\cellcolor[HTML]{000000}} \color[HTML]{F1F1F1} {\cellcolor{white}} \color{black} -- & {\cellcolor[HTML]{000000}} \color[HTML]{F1F1F1} {\cellcolor{white}} \color{black} -- & {\cellcolor[HTML]{000000}} \color[HTML]{F1F1F1} {\cellcolor{white}} \color{black} -- & {\cellcolor[HTML]{000000}} \color[HTML]{F1F1F1} {\cellcolor{white}} \color{black} -- & {\cellcolor[HTML]{000000}} \color[HTML]{F1F1F1} {\cellcolor{white}} \color{black} -- & {\cellcolor[HTML]{000000}} \color[HTML]{F1F1F1} {\cellcolor{white}} \color{black} -- \\
\end{tabular}
}